\documentclass[1p,times]{elsarticle}

\usepackage{natbib}
\usepackage{amssymb}
\usepackage{amsmath}
\usepackage{bm}
\usepackage{xcolor}

\usepackage{tikz}
\usepackage{pgfplots}
\pgfplotsset{compat=1.18}
\usepgfplotslibrary{fillbetween}
\usetikzlibrary{shapes,arrows,positioning,calc}

\tikzstyle{block} = [draw, rectangle, 
    minimum height=3em, minimum width=6em]
\tikzstyle{block_noline} = [rectangle, 
    minimum height=3em, minimum width=6em]
\tikzstyle{input} = [coordinate]
\tikzstyle{output} = [coordinate]
\tikzstyle{pinstyle} = [pin edge={to-,thin,black}]

\makeatletter
\def\blfootnote{\xdef\@thefnmark{}\@footnotetext}
\makeatother

\usepackage{blox}

\usepackage{caption}
\usepackage{subcaption}

\usepackage{microtype}

\usepackage{algorithm}
\usepackage{algpseudocode}

\journal{Signal Processing}

\usepackage[acronym]{glossaries}
\newacronym{new}{AGNesF}{adaptive Gaussian nested filter}
\newacronym{AWGN}{AWGN}{addtive white {G}aussian noise}
\newacronym{apf}{APF}{auxiliary particle filter}
\newacronym{aesmc}{AESMC}{auto-encoding sequential Monte Carlo}

\newacronym{bpf}{BPF}{bootstrap particle filter}
\newacronym{ckf}{CKF}{cubature Kalman filter}

\newacronym{dpf}{DPF}{differentiable particle filter}
\newacronym{ekf}{EKF}{extended Kalman filter}
\newacronym{enkf}{EnKF}{ensemble Kalman filter}
\newacronym{ess}{ESS}{effective sample size}

\newacronym{fvo}{FIVOs}{filtering variational objectives}
\newacronym{fcnn}{FCNN}{fully connected neural network}

\newacronym{gnn}{GNN}{generative neural network}

\newacronym{hmm}{HMM}{heterogeneous multi-scale model}

\newacronym{iid}{i.i.d.}{independent and identically distributed}
\newacronym{ipf}{IPF}{island particle filter}
\newacronym{iapf}{IAPF}{improved auxiliary particle filter}

\newacronym{kf}{KF}{Kalman filter}

\newacronym{mse}{MSE}{mean square error}

\newacronym{nhf}{NHF}{nested hybrid filter}
\newacronym{ngf}{NGF}{nested Gaussian filter}
\newacronym{npf}{NPF}{nested particle filter}
\newacronym{nsmc}{NSMC}{nested sequential Monte Carlo}
\newacronym{nmse}{NMSE}{normalized mean square error}
\newacronym{pdf}{pdf}{probability density function}
\newacronym{pmcmc}{PMCMC}{particle Markov chain Monte Carlo}
\newacronym{pf}{PF}{particle filter}
\newacronym{pl}{PL}{particle learning}
\newacronym{pde}{PDE}{partial differential equation}

\newacronym{qmc}{QMC}{quasi Monte Carlo}
\newacronym{qkf}{QKF}{quadrature Kalman filter}

\newacronym{rv}{r.v.}{random variable}
\newacronym{rml}{RML}{recursive maximum likelihood}
\newacronym{rnn}{RNN}{recurrent neural network}

\newacronym{sir}{SIR}{sequential importance resampling}
\newacronym{smc}{SMC}{sequential Monte Carlo}
\newacronym{smc2}{SMC$^2$}{sequential Monte Carlo square}
\newacronym{sqmc}{SQMC}{sequential quasi-Monte Carlo}
\newacronym{sde}{SDE}{stochastic differential equation}
\newacronym{stpf}{ST-PF}{space-time particle filter}
\newacronym{ssm}{SSM}{state-space model}
\newacronym{ukf}{UKF}{unscented Kalman filter}
\newacronym{ut}{UT}{unscented transform}

\newacronym{vsmc}{VSMC}{variational sequential Monte Carlo}

\newacronym{wrt}{w.r.t.}{with respect to}

\def\b{{\mathbf b}}
\def\c{{\mathbf c}}

\def\x{{\mathbf x}}
\def\y{{\mathbf y}}
\def\z{{\mathbf z}}

\def\A{{\mathbf A}}
\def\C{{\mathbf C}}

\newcommand{\reals}{\mathbb R}      

\newcommand{\mname}{StateMixNN}

\begin{document}
    
    \begin{frontmatter}
    
       
    
    \title{Learning state and proposal dynamics in state-space models using differentiable particle filters and neural networks} 
    
    \author[label1]{Benjamin Cox}
    \author[label2]{Santiago Segarra}
    \author[label1]{V\'{i}ctor Elvira}
    \affiliation[label1]{organization={School of Mathematics, University of Edinburgh},
                addressline={James Clerk Maxwell Building},
                city={Edinburgh},
                postcode={EH9 3FD}, 
                state={Scotland},
                country={United Kingdom}}
    \affiliation[label2]{organization={Electrical and Computing Engineering, Rice University},
                addressline={6100 Main Street MS 366},
                city={Houston},
                postcode={77005},
                state={Texas},
                country={USA}}
    
    
    \begin{abstract}
    State-space models are a popular statistical framework for analysing sequential data.
    Within this framework, particle filters are often used to perform inference on non-linear state-space models.
    We introduce a new method, \mname, that uses a pair of neural networks to learn the proposal distribution and transition kernel of a particle filter. 
    Both distributions are approximated using multivariate Gaussian mixtures.
    The component means and covariances of these mixtures are learnt as outputs of learned functions.
    Our method is trained targeting the log-likelihood, thereby requiring only the observation series, and combines the interpretability of state-space models with the flexibility and approximation power of artificial neural networks.
    The proposed method significantly improves recovery of the hidden state in comparison with the state-of-the-art, showing greater improvement in highly non-linear scenarios.
    \end{abstract}
     
    
    
    \blfootnote{B.C. acknowledges support from the \emph{Natural Environment Research Council} of the UK through a SENSE CDT studentship (NE/T00939X/1). The work of V. E. is supported by the \emph{Agence Nationale de la Recherche} of France under PISCES (ANR-17-CE40-0031-01), the Leverhulme Research Fellowship (RF-2021-593), and by ARL/ARO under grant W911NF-22-1-0235. In addition, we acknowledge support from the 2022 University of Edinburgh – Rice University Strategic Collaboration Award.}
    
    \end{frontmatter}
    
    \section{Introduction}
        \label{sec:intro}

        In many fields of science and engineering, it is common to process sequential data resulting from the observation of dynamical systems.
        These systems appear in fields such as target tracking, \citep{wang2017survey}, finance \citep{virbickaite2019particle}, epidemiology \citep{patterson2017statistical}, ecology \citep{newman2023state}, and meteorology \citep{clayton2013operational}.
        We can describe these systems and their observations mathematically, utilising the \gls{ssm} framework.
        \glspl{ssm} represent a dynamical system via a latent state and a series of related, noisy, observations, encoding the system via a pair of distributions describing the latent state dynamics and its relationship to the observation.
        Within state-space modelling, it is common to compute the distribution of the state conditional on the distribution of the previous state and the current observation, a problem known as the filtering problem.
        
        In the case of linear dynamics with Gaussian noise, the \gls{kf} provides the optimal solution to the filtering problem via a sequence of closed form estimates.
        However, if the system is more complex, the \gls{kf} can no longer be used directly.
        A number of extensions to the \gls{kf} to non-linear systems have been proposed, such as the \gls{ekf} \citep{ribeiro2004kalman} and the \gls{ukf} \citep{wan2000unscented}.
        These methods approximate the state posterior via a Gaussian distribution, which is not always appropriate.
        The \gls{pf}, also known as \gls{smc}, is an alternative method for the general \gls{ssm}, which approximates the state posterior via a series of Monte Carlo samples \citep{Gordon93, djuric2003particle, doucet2009tutorial}.

        {In \glspl{pf}, we must evaluate the likelihood of the distribution of the state given the previous state, known as the transition kernel.}
        If the form of the transition kernel is known, with only the parameters unknown, then several methods exist to estimate the system parameters, and hence estimate the transition kernel \citep[Chapter 12]{sarkka2013bayesian}.
        Furthermore, learning parameters in the \gls{pf} has been greatly eased with the advances in \glspl{dpf} \citep{corenflos2021differentiable,scibior2021differentiable,chen2023overview,li2023differentiable}, which allow the use of gradient-based optimisation methods for parameter estimation.

        Differentiable particle filters (DPFs) modify the resampling step of the \gls{pf} to be differentiable, and therefore, when combined with differentiable transition and observation likelihoods, makes the overall particle filter differentiable. 
        There exist a number of differentiable particle filters, which utilise many different methods for making the resampling step differentiable, such as soft resampling \citep{karkus2018particle}, stop-gradient operators \citep{scibior2021differentiable}, or optimal transport schemes \citep{corenflos2021differentiable}.

        If the form of the transition kernel is unknown, current methods require one to be assumed.
        {In practice, it is often the case that the transition kernel has unknown form, such as when the underlying dynamics are not known.
        In these cases, it is required to perform filtering without prior knowledge of the transition kernel, and therefore the form and parameters of the transition kernel must be estimated.}
        In addition to the transition kernel, in the \gls{pf} we must provide a proposal distribution to generate Monte Carlo samples.
        The diversity of the samples, which is critically important to the method's ability to represent the system, is heavily dependent on the proposal distribution.
        
        One approach to choosing the proposal distribution is to use the \gls{bpf} \citep{Gordon93}, which uses the state transition kernel as the proposal.
        However, the bootstrap proposal does not incorporate the observation, and therefore does not utilise all available information at each time step.
        Not using observation information can cause significant issues, for example, if the state transition is a very diffuse distribution but the observation is very informative, few, if any, particles will have high posterior probability, and therefore the effective sample size will be very low.
        Several methods incorporate the observation in their proposal distribution, such as the \gls{apf} \citep{pitt1999filtering,elvira2019elucidating,branchini2021optimized,elvira2018improved}, which adjusts the particle weights based on both the transition dynamics and the likelihood of the new observation.
        Deep learning methods have also been used within the \gls{smc} and \gls{pf} framework to learn the proposal distribution, such as 
        neural adaptive \gls{smc} \citep{gu2015neural} and the variational \gls{smc} \citep{naesseth2018variational}.
        
        \vspace{1em}\noindent{\textbf{Contribution}.} In this paper, we propose \mname, a method to approximate the transition kernel and proposal distribution of a particle filter using a pair of adaptive Gaussian mixtures.\footnote{A limited version of this work was presented by the authors in the conference paper \cite{cox2024end}, which presented a version of this work learning only the proposal distribution. 
        We propose here a more advanced method that can learn both the proposal distribution and the transition kernel. 
        We include here extensive methodological discussion, and provide practical guidance for practitioners. 
        Finally, we include a large number of numerical experiments, which were not present in the prior work.}
        \mname~improves upon methods that learn only the transition dynamics, as we also estimate the optimal proposal density, and therefore incorporate more information in our estimates than would otherwise be possible.
        
        \mname~learns the mean and covariance parameters of the components of a multivariate Gaussian mixture as the output of a dense neural network. 
        By estimating both the transition and proposal distributions, we can estimate the hidden state from an observation series given only the observation model, which cannot be estimated here for identifiability reasons.
        This approximation allows us to estimate distributions resulting from complex models, becoming more expressive as the number of mixture components increases.
        
        In order to train the neural networks, we utilise the \gls{dpf} framework of \citet{scibior2021differentiable}, allowing gradient propagation through the resampling step of the particle filter, and therefore the use of gradient-based optimisation schemes.
        \mname~is semi-supervised since it trains targeting the parameter log-likelihood, which requires the sequence of observations, but does not require the underlying hidden states.

        {This paper is laid out as follows.
        First, we outline the requisite background knowledge in Section~\ref{sec:background}.
        Next, we construct our proposed algorithm,~\mname, in Section~\ref{sec:methodology}.
        We then discuss and justify the choices made constructing~\mname~in Section~\ref{sec:discuss}.
        In Section~\ref{sec:experiments}, we perform numerical experiments showcasing~\mname~against several existing methods.
        Finally, we conclude with Section~\ref{sec:conclusions}, highlighting the key points of the paper.}

    \section{Background}
        \label{sec:background}

    \subsection{State-space models (SSMs)}
        We are interested in state-space dynamical systems, which we can describe by
            \begin{align}
                \begin{split}
                    \x_t &\sim f(\x_t | \x_{t-1}; \bm\theta^{(f)}),\\
                    \y_t &\sim g(\y_t | \x_t; \bm\theta^{(g)}),
                \end{split} \label{eq:ssm}
            \end{align}
        where $t \in \{1,\dots, T\}$ denotes discrete time, $\x_t \in \reals^{d_x}$ is the state of the system at time $t$, $\y_t \in \reals^{d_y}$ is the observation associated with $\x_t$, $\bm\theta^{(g)}$ and $\bm\theta^{(f)}$ are sets of parameters relating to the observation and state dynamics respectively, and the distributions $f$ and $g$ encode the transition and observation model respectively.
        
        In terms of probability densities, $f(\x_t | \x_{t-1}; \bm\theta^{(f)})$ is the conditional density of the state $\x_t$ given $\x_{t-1}$, and $g(\y_t | \x_{t}; \bm\theta^{(g)})$ is the conditional density of the observation $\y_t$ given the hidden state $\x_t$.
        The initial value of the state, $\x_0$, is distributed as $\x_0 \sim p(\x_0|\bm\theta^{(p)})$.
        The hidden state sequence $\x_{0:T}$ is not observed, while the related sequence of observations $\y_{1:T}$ is observed.
        In many cases, we wish to estimate the sequence of hidden states given the observation sequence. 
        If we estimate the state at time $t$, given by $\x_t$, conditional on observations collected up to and including time $t$, given by $\y_{1:t}$, then this we call this the filtering problem, with $p(\x_t|\y_{1:t})$ being known as the filtering distribution.

    \subsection{Particle filtering}
    \label{sec:pf_methods}
        Filtering methods for \glspl{ssm} aim to probabilistically recover the state of the systems described in Eq.~\eqref{eq:ssm} by estimating the posterior \gls{pdf} of the state $\x_t$ conditional on all prior observations $\y_{1:t}$, i.e., $p(\x_t | \y_{1:t}; \bm\theta^{(f)})$. 
        \glspl{pf} approximate this \gls{pdf} utilising a set of $K$ Monte Carlo samples and associated weights, $\{ \x_t^{(k)}, \tilde{w}_t^{(k)} \}_{k=1}^K$. 
        The filtering distribution can thus be approximated by
            \begin{equation}
                p(\x_t | \y_{1:t}; \bm\theta^{(f)}) \approx \sum_{k=1}^K \tilde{w}_t^{(k)} \delta_{\x_t^{(k)}}.
            \end{equation}

        A ubiquitous algorithm for particle filtering is the \gls{sir} algorithm, which we give in Alg.~\ref{alg:sir_pf}. 
        In this method, at time step $t$, we compute a set of particles and associated importance weights via the following steps.
        {First, we draw $K$ ancestor indices $\{a_{t}^{(k)}\}_{k=1}^{K}$, (line \ref{step_pf:ancestor}), which is equivalent to resampling the particle set $K$ times with replacement, with the probability of drawing particle $\x_{t-1}^{(k)}$ equal to its weight $\bar{w}_{t-1}^{(k)}$.
        Resampling is important to avoid weight degeneracy in the filter, as it helps to avoid the weights becoming concentrated in a diminutive subset of the sampled particles. 
        However, resampling results in path degeneracy \cite{elvira2017improving}, where all particles will eventually share a common ancestor trajectory.
        Path degeneracy does not impact the properties of the filter at each time step however, so is considered far less damaging than weight degeneracy.
        Next, we draw $K$ particles $\{\x_t^{(k)}\}_{k=1}^{K}$ from the proposal distribution $\pi(\x_t|\x_{t-1}^{a_{t}^{(k)}}, \y_{t}; \bm\theta^{(\pi)})$ (line \ref{step_pf:draw}). 
        Then, we compute the importance weights $\{{w}_t^{(k)}\}_{k=1}^{K}$ (line \ref{step_pf:weights}), which takes into account the transition kernel and the current observation.
        Finally, we normalise the weights to sum up to $1$, obtaining $\{\bar{w}_t^{(k)}\}_{k=1}^{K}$ (line \ref{step_pf:normweights}).}
        
            {\footnotesize
            \begin{algorithm}
            {
                \caption{Sequential importance resampling (SIR)}\label{alg:sir_pf}
                \begin{algorithmic}[1]
                    \State Draw $\x_0^{(k)} \sim p(\x_0|\bm\theta^{(p)})$, for $k \in 1,\dots,K$.
                    \State Set $\tilde{w}_0^{(k)} = 1/K$, for $k \in 1,\dots,K$.
                    \For{$t \in 1,\dots,T$ and $k \in 1,\dots,K$}
                        \State Sample $a_{t}^{(k)} \sim \mathrm{Categorical}(\bar{w}_{t-1})$. \label{step_pf:ancestor}
                        \State Set $\tilde{w}_t^{(k)} = \frac{1}{K}$.   
                        \State Sample $\x_t^{(k)} \sim \pi(\x_t|\x_{t-1}^{a_{t}^{(k)}}, \y_{t}; \bm\theta^{(\pi)})$. \label{step_pf:draw}
                        \State Compute $w_t^{(k)} = \frac{g(\y_t|\x_t^{(k)}; \bm\theta^{(g)})f(\x_t^{(k)}|\x_{t-1}^{a_{t}^{(k)}}; \bm\theta^{(f)})}{\pi(\x_t|\x_{t-1}^{a_{t}^{(k)}}, \y_{t}; \bm\theta^{(\pi)})}$. \label{step_pf:weights}
                        \State Compute $\bar{w}_t^{(k)} = \tilde{w}_{t-1}^{(k)}w_t^{(k)}/\sum_{k=1}^{K}\tilde{w}_{t-1}^{(k)}w_t^{(k)}$. \label{step_pf:normweights}   
                    \EndFor
                \end{algorithmic}
                }
            \end{algorithm}
            }

        Given the \gls{sir} algorithm, it is apparent that the choice of proposal distribution $\pi(\x_t|\x_{t-1}, \y_{t}; \bm\theta^{(\pi)})$ is critical to the success of the estimation; the particles should be concentrated in regions of the state space with high probability mass.
        There exist several methods to choose the proposal distribution.
        
        One such method is the \gls{bpf} \citep{Gordon93}, where the proposal distribution is set equal to the transition kernel of the \gls{ssm}, with $\pi(\x_t|\x_{t-1},\y_t; \bm\theta^{(\pi)})=f(\x_t|\x_{t-1}; \bm\theta^{(f)})$.
        Such an approach is simple, and results in less computation due to cancellation in line \ref{step_pf:weights} of Alg.~\ref{alg:sir_pf}, resulting in $w_t^{(k)} = g(\y_t|\x_t^{(k)} \bm\theta^{(g)})$ for the bootstrap particle filter.
        However, the \gls{bpf} does not incorporate the observation $\y_t$ in the proposal distribution, and therefore omits potentially important information.
        For example, if the transition kernel is far more diffuse than the observation distribution, then the observation contains a lot of information relative to the previous state, which the \gls{bpf} does not use.
        
        The optimal proposal distribution incorporates the observation at the current time step $t$, i.e., $\pi(\x_t|\x_{t-1},\y_t; \bm\theta^{(\pi)})=p(\x_t|\x_{t-1},\y_t)$. 
        In general distribution is intractable or unknown.
        Methods such as the \gls{apf} and its derivatives \citep{pitt1999filtering,elvira2018improved,elvira2019elucidating,branchini2021optimized} incorporate the observation information via a pre-weighting step.
        
        \citet{gama2022unrolling, gama2023unsupervised} propose several methods to learn the proposal distribution in a general setting, such as modelling it as a Gaussian distribution with mean and covariance output by a neural network, or utilising normalising flows.
        These methods learn a distinct network for each time step, and therefore do not generalise outside of a single series of observations.
        
    \subsection{Differentiable particle filters}
    \label{sec:diff_pf}
            following
        The particle filter as given in Alg.~\ref{alg:sir_pf} {contains non-differentiable operations using} the weights and particle values, and hence is not differentiable {with respect to the model parameters $\bm\theta^{(f)}, \bm\theta^{(\pi)},$ or $\bm\theta^{(g)}$},  from the two sampling steps (lines \ref{step_pf:draw} and \ref{step_pf:ancestor}).
        {The proposal sampling step (line \ref{step_pf:draw}) can often be rewritten to be differentiable with respect to the parameters $\bm\theta^{(f)}$ and $\bm\theta^{(\pi)}$ using distribution specific reparametrisations, such as the reparametrisation trick for Gaussian distributions\footnote{The Gaussian distribution $\mathcal{N}(\bm\mu, \bm\Sigma)$ can be sampled differentiably by noting that $\mathbf{x}\sim\mathcal{N}(\bm\mu, \bm\Sigma)\sim\bm\mu + \bm L \bm\varepsilon, $ where $\bm\Sigma = LL^T$, and $\bm\varepsilon \sim \mathcal{N}(\bm0, \bm 1)$, and taking gradients of this transformation accordingly, noting that $\bm\varepsilon$ is independent of the parameters of the initial distribution, and is therefore a constant for the purposes of the derivative. } \citep{kingma2013auto}.}
        
        {The resampling step (line \ref{step_pf:ancestor}) remains, which requires sampling a categorical or multinomial distribution, depending on implementation.
        Sampling either of these distributions is a non differentiable operation, as infinitesimal changes in the weights result in discrete changes in the sampled indices \citep{zhu2020towards}.
        A heuristic reasoning for this non-differentiability can be made by noting that $a \sim \mathrm{Categorical}(p)$ can be sampled by drawing $u \sim \mathrm{Uniform}(0, 1)$, and then setting $a = \max N \ \mathrm{s.t.} \sum_{n=1}^{N}p_n \leq u$.
        If we have $\sum_{n=1}^{N}p_n = u$ for some $I$, then an infinitesimal increase in $p_N$ by any $\epsilon > 0$, will result in a discrete change in $a$ from $N$ to $N+1$.
        Therefore, sampling $\mathrm{Categorical}(p)$ is not differentiable with respect to $p$.}
        
        Due to resampling not being differentiable, any function of the weights is non-differentiable, with one such function being the posterior likelihood of the parameters $\ell(\bm\theta|\y_{1:T})$, which is a common target function for parameter estimation in particle filtering.
        Note that the particle trajectories are also functions of the weights, so functions of trajectories are also non-differentiable if using Alg.~\ref{alg:sir_pf}.
        
        Differentiable particle filtering (DPF) modifies the resampling step to achieve differentiability.
        Therefore, when combined with a differentiable transition step, the overall particle filter is differentiable \citep{corenflos2021differentiable,chen2023overview,scibior2021differentiable,li2023differentiable}.
        {There exist several DPF methods, each utilising a different method to make the resampling step differentiable.
        In this work, we use the stop-gradient DPF of \citep{scibior2021differentiable}, which we present in Alg.~\ref{alg:sgd_dpf}.
        The stop-gradient DPF yields gradient estimates of a given loss function with minimal computational overhead \citep{scibior2021differentiable}.
        This method achieves differentiability using two steps.
        First, the stop-gradient operator is applied to the weights when sampling particle ancestry in line \ref{step_dpf:resample}, and thereby we do not require to compute gradients through the discrete sampling.
        However, gradient information is then propagated through a transformation of the weights in line \ref{step_dpf:weight_transform}.
        Therefore, we can take gradients of filter outputs, such as the particle trajectories and the particle weights, with respect to the filter inputs.
        An overview of differentiable particle filter methods is given in \cite{chen2023overview}, and a detailed treatment of the specific filter we use is given in \cite{scibior2021differentiable}.}
        
        \begin{algorithm}[ht]
            \small
            \caption{Stop-gradient differentiable particle filter (DPF) \hfill \citep{scibior2021differentiable} \hspace{1em} \null}
            \label{alg:sgd_dpf}
            \begin{algorithmic}[1]
                \State Draw $\x_0^{(k)} \sim p(\x_0|\bm\theta^{(p)})$, for $k \in 1,\dots,K$.
                \State Set $\tilde{w}_0^{(k)} = 1/K$, for $k \in 1,\dots,K$.
                \For{$t \in 1,\dots,T$ and $k \in 1,\dots,K$}
                    \State Sample $a_{t}^{(k)} \sim \mathrm{Categorical}(\bot(\bar{w}_{t-1}))$. \label{step_dpf:resample}
                    \State Set $\tilde{w}_t^{(k)} = \frac{1}{K} \ \bar{w}_t^{a_{t}^{(k)}} / \bot(\bar{w}_t^{a_{t}^{(k)}})$. \label{step_dpf:weight_transform}
                    \State Sample $\x_t^{(k)} \sim \pi(\x_t|\x_{t-1}^{a_{t}^{(k)}}, \y_{t}; \bm\theta^{(\pi)})$. \label{step_dpf:draw}
                    \State Compute $w_t^{(k)} = \frac{g(\y_t|\x_t^{(k)}; \bm\theta^{(g)})f(\x_t^{(k)}|\x_{t-1}^{a_{t}^{(k)}}; \bm\theta^{(f)})}{\pi(\x_t|\x_{t-1}^{a_{t}^{(k)}}, \y_{t}; \bm\theta^{(\pi)})}$. \label{step_dpf:weights}
                    \State Compute $\bar{w}_t^{(k)} = \tilde{w}_{t-1}^{(k)}w_t^{(k)}/\sum_{k=1}^{K}\tilde{w}_{t-1}^{(k)}w_t^{(k)}$. \label{step_dpf:normweights}   
                \EndFor
            \end{algorithmic}
        \end{algorithm}
            
        We can use differentiable particle filters to estimate the gradient of the parameter log-likelihood, and therefore apply gradient methods to estimate unknown parameters of our state-space model.
        However, as the log-likelihood is estimated, we will also have noise in gradients thereof, and therefore the parameter estimation scheme must be robust to noisy gradients.
        Noisy gradients are common in deep learning, as it is typical to compute the gradient of a stochastically selected subset of the training data, which results in gradient noise due to sampling. 
        Several optimisation methods have been proposed for use in deep learning, with a common feature being robustness to stochastic gradients.
        We can utilise these methods when estimating parameters using differentiable particle filters, with schemes such as ADAM \citep{kingma2015method}, RADAM \citep{liu2019variance}, and Novograd \citep{ginsburg2019training, li2019jasper}, all being applicable to this problem.
    
    \section{Proposed algorithm}
        \label{sec:methodology}
        We propose \mname, a method to learn the transition and proposal distribution of a generic state-space model.
       {We approximate the transition kernel and the proposal distribution by a pair of multivariate Gaussian mixtures parametrised by neural networks.}

        By combining learnt estimates of both the transition kernel and the proposal distribution, we can learn a generic state-space model conditional only on knowledge of the observation model. 
        {Note that we discuss and justify several of the choices made here, such as the use of Gaussian mixtures, equal mixture weights, and diagonal covariances, in Section~\ref{sec:discuss}.}

    \subsection{Parametrising the Gaussian mixture}

        \mname~learns the transition kernel conditional on only the previous state value, thus preserving the Markovianity of the SSM.
        We approximate $f(\x_t|\x_{t-1}; \bm\theta^{(f)})$ by an equally weighted multivariate Gaussian mixture of $S_f$ components with diagonal covariances
            \begin{equation}
                \label{eq:state_dist}
                f(\x_t|\x_{t-1}; \bm\theta^{(f)}) := (S^{(f)})^{-1}\sum_{s=1}^{S_f}f_s(\x_t|\x_{t-1}; \bm\theta^{(f)}),
            \end{equation}
        with $s \in \{1, \dots, S_f\}$, where $\bm\theta^{(f)}$ are the parameters of the state neural network, and $f_s$ is given by
        \begin{equation}
            f_s(\x_t|\x_{t-1}; \bm\theta^{(f)}) = \mathcal{N}\left(\mu^{(f)}_s(\x_{t-1}), \Sigma^{(f)}_s(\x_{t-1})\right).
        \end{equation}

        \mname~learns the proposal distribution conditioned on the previous state value and the current observation, as is the case for the intractable optimal proposal distribution.
        We parametrise $\pi(\x_t|\x_{t-1}, \y_{t}; \bm\theta^{(\pi)})$ as an equally weighted mixture of $S_\pi$ multivariate Gaussian distributions with diagonal covariances
            \begin{equation}
                \label{eq:prop_dist}
                \pi(\x_t|\x_{t-1}, \y_{t}; \bm\theta^{(\pi)}) := (S^{(\pi)})^{-1}\sum_{s=1}^{S_\pi}\pi_s(\x_t|\x_{t-1}, \y_{t}; \bm\theta^{(\pi)}),
            \end{equation}
        with $s \in \{1, \dots, S_\pi\}$, where $\bm\theta^{(\pi)}$ are the parameters of the proposal neural network, and $\pi_s$ is given by
        \begin{equation}
                \pi_s(\x_t|\x_{t-1}, \y_{t}; \bm\theta^{(\pi)}) = \mathcal{N}\left(\mu^{(\pi)}_s(\x_{t-1}, \y_t), \Sigma^{(\pi)}_s(\x_{t-1}, \y_t)\right).
        \end{equation}
            
        Mixtures of multivariate Gaussians can represent a wide range of possible distributions, offering flexibility and expressiveness that is beneficial for modelling complex systems \citep{bugallo2017adaptive}.

    \subsection{Network architecture and learning}
    \label{sec:arch}
        {For real data, the transition kernel of a state-space model is, in general, unknown.
        By estimating the transition kernel, we can utilise the state-space model framework to perform estimation in an unknown model scenario.
        Furthermore, the optimal proposal distribution is separate from the state-space model, and is in general intractable.
        Using a proposal distribution that approximates the optimal proposal distribution will improve the performance of the particle filter by way of making the sampling more efficient, thereby yielding more accurate estimates of state quantities by way of increased effective sample size.}

        {The transition kernel can be approximated utilising neural networks, which are efficient to evaluate, and can approximate any general continuous function (in the infinite width/depth limit) due to the universal approximation theorem.
        Small neural networks are simple to fit via gradient descent, and easy to implement due to the proliferation of deep learning frameworks.
        In this work we restrict ourselves to multilayer perceptrons, however the method could be extended to other forms of network.}
        
        Our method learns the mean functions $\mu^{(f)}(\cdot), \ \mu^{(\pi)}(\cdot)$ and the covariance functions $\Sigma^{(f)}(\cdot), \ \Sigma^{(\pi)}(\cdot)$ for the transition and proposal distributions using a pair of dense neural networks. 
        We denote these networks by $\mathrm{NN}^{(f)}(\x_{t-1}; \bm\theta^{(f)})$ for the network associated with the transition kernel and by $\mathrm{NN}^{(\pi)}(\x_{t-1}, \y_t; \bm\theta^{(\pi)})$ for the network associated with the proposal distribution.
        
        The transition network takes as input only the previous particle value, therefore preserving the Markovianity assumption of the \gls{ssm}, which is often violated when using neural networks \citep{medsker1999recurrent, schuster1997bidirectional, salehinejad2017recent, zhou2021informer}.
        The network $\mathrm{NN}(\cdot; \bm\theta)$ is comprised of $L$ layers, with
            \begin{equation}
                \label{eq:nn_setup}
                \mathrm{NN}(\cdot; \bm\theta) = \z_{L}, \qquad \z_l = \rho_l(\A_l\z_{l-1} + \b_l),
            \end{equation}
        for $l \in 1,\dots,L$. 
        Therefore, $\bm\theta = \{\A_1,\b_1,\dots,\A_L,\b_L\}$ are the learned parameters for a network.
        For the proposal network the initial value $\z_0^{(\pi)} = [\x_{t-1}^\intercal,\y_t^\intercal]^\intercal$ is the concatenation of the previous state $\x_{t-1}$ and the current observation $\y_t$.
        Thus, $\z_0^{(\pi)}$ has dimension $d_0^{(\pi)} = d_x + d_y$. 
        For the state network the initial value $\z_0^{(f)} = \x_{t-1}$ is the previous state $\x_{t-1}$.
        Thus, $\z_0^{(f)}$ has dimension $d_0^{(f)} = d_x$. 

        For both networks, the dimension of the output
            \begin{equation}
                \label{eq:output_layer}
                \z_{L}=[\bm\mu^{(1)\intercal},\c^{(1)\intercal},\dots,\bm\mu^{(S)\intercal},\c^{(S)\intercal}]^\intercal
            \end{equation} 
        is $d_L := 2Sd_x$, as for each of the mixture components of both distributions, we require a $d_x$-dimensional mean vector $\bm\mu^{(n)}$ and a $d_x$-dimensional covariance scale vector $\c^{(n)}$.

        In both cases, we construct a Gaussian mixture distribution from a neural network $\mathrm{NN}(\cdot;\bm\theta)$ by
            \begin{equation}
                \label{eq:makedist}
                \mathcal{GM}(\mathrm{NN}(\cdot;\bm\theta)) = \sum_{s=1}^S S^{-1}\mathcal{N}(\bm\mu^{(s)},\C^{(s)}),
            \end{equation}
        where $S$ is the number of mixture components, with $\bm\mu^{(s)}, \bm c^{(s)}, s\in1,\dots,S$ extracted from $\z_L$ by indexing per Eq.~\eqref{eq:output_layer}, where we discard the components of $\bm\theta$ that are not relevant to $\mathrm{NN}$, and $\C^{(s)} = \mathrm{diag}(\c^{(s)})^2$. 
        For example, when constructing $f$, we use $\mathcal{GM}(\mathrm{NN}^{(f)}(\cdot;[\bm\theta^{(f)}, \bm\theta^{(\pi)}]))$, discarding $\bm\theta^{(\pi)}$ it is not used in $\mathrm{NN}^{(f)}$. 
        We make use of this property in Alg.~\ref{alg:method_update_step} and Fig.~\ref{fig:update_bd}.
        The learned parameters for a neural network are the weights $\A_l$ and the biases $\b_l$, with $l \in 1,\dots,L$.
        The number of layers $L$, the dimension of each layer $d_l, l \in 1,\dots,L$, and the activation functions $\rho_l, l \in 1,\dots,L$ are fixed as part of the network architecture.
        
        We learn separate networks for the transition and proposal distributions, denoted by $\mathrm{NN}^{(f)}$ and $\mathrm{NN}^{(\pi)}$ respectively.
        Therefore, when learning the transition kernel we estimate $\bm\theta^{(f)} := \{\A_1^{(f)}, \b_1^{(f)}, \dots, \A_{L^{(f)}}^{(f)}, \b_{L^{(f)}}^{(f)}\}$, and estimate $\bm\theta^{(\pi)} := \{\A_1^{(\pi)}, \b_1^{(\pi)}, \dots, \A_{L^{(\pi)}}^{(\pi)}, \b_{L^{(\pi)}}^{(\pi)}\}$ when learning the proposal distribution.

        To train the networks we maximise the log-likelihood, given by 
            \begin{equation}
                \label{eq:loglik}
                \ell(\bm\theta|\y_{1:T}) \propto \sum_{t=1}^{T} \left(\sum_{k=1}^{K} \log\left(w_t^{(k)} \cdot \tilde{w}_{t-1}^{(k)}\right)\right),
            \end{equation}
        {where $w_t^{(k)}$ and $\tilde{w}_{t-1}^{(k)}$ are the unnormalised and normalised weights of the \gls{pf} in Alg.~\ref{alg:sir_pf}, and we assume a uniform flat improper prior $p(\bm\theta) \propto 1$ \citep[Chapter 12]{sarkka2013bayesian}.
        Equation~\eqref{eq:loglik} is a stochastic estimate of the log-likelihood, which is generally not available in closed form. 
        In order to maximise the likelihood of our learned parameters, we aim to maximise Equation~\eqref{eq:loglik}.}
        
        Note that in the case that we perform resampling at every step, Eq.~\eqref{eq:loglik} reduces to
            \begin{equation}
                \label{eq:loglik_resampleevery}
                \ell(\bm\theta|\y_{1:T}) \propto \sum_{t=1}^{T} \left(\sum_{k=1}^{K} \log\left(w_t^{(k)}\right)\right),
            \end{equation}
        as the previous step normalised weights are identically equal to $1/K$, {and thus reduce to a constant factor which can be ignored due to weight normalisation}.
        Note that the weights are dependent on the parameters $\bm\theta^{(f)}, \bm\theta^{(g)}$, and $\bm\theta^{(\pi)}$, as the weights are computed using densities dependent on these parameters.
        
        In our case, we assume that the observation model $g$ is known, and hence we omit the dependence on $\bm\theta^{(g)}$.
        The log-likelihood is maximised when all weights are equal, a desirable property as it reduces weight degeneracy over time \citep{sarkka2013bayesian}.
        Furthermore, maximising the log-likelihood does not require knowledge of the true value of the hidden state, which is often unavailable, requiring only the observation series.
 
    \subsection{\mname~algorithm}
        {\footnotesize
        \begin{algorithm}[H]
            \caption{\mname~algorithm \hfill $\mathrm{\mname}(B, J, A, \y, \mathrm{NN}^{(f)}, \mathrm{NN}^{(\pi)})$ \hspace{1em} \null}\label{alg:method_outer_alg}
            \begin{algorithmic}[1]
                \State \textbf{Input:} Number of batches $B$, steps per batch $J$, number of iterations $A$, observations $\y$, transition network $\mathrm{NN}^{(f)}$, proposal network $\mathrm{NN}^{(\pi)}$.
                \State Initialise $\bm\theta_{0}^{(\pi)} := \{\A_1^{(\pi)},\b_1^{(\pi)},\dots,\A^{(\pi)}_{L^{(\pi)}},\b^{(\pi)}_{L^{(\pi)}}\}$. \label{step:smnn:init_prop}
                \State Initialise $\bm\theta_{-1}^{(f)} := \{\A_1^{(f)},\b_1^{(f)},\dots,\A^{(f)}_{L^{(f)}},\b^{(f)}_{L^{(f)}}\}$. \label{step:smnn:init_trans}
                \State Set $\bm\theta_{0}^{(f)} := \mathrm{ConditionalUpdate}(B, J, \bm\theta_{-1}^{(f)}, \bm\theta_{-1}^{(f)}, \y)$ {using Alg.~\ref{alg:method_train_alg}} \label{step:smnn:bs_trans}
                \For{$a \in 1,\dots,A$}
                    \State Set $\bm\theta_{a}^{(\pi)} := \mathrm{ConditionalUpdate}(B, J, \bm\theta_{a-1}^{(\pi)}, \bm\theta_{a-1}^{(f)}, \y)$ {using Alg.~\ref{alg:method_train_alg}}. \label{step:smnn:prop_ud}
                    \State Set $\bm\theta_{a}^{(f)} := \mathrm{ConditionalUpdate}(B, J, \bm\theta_{a-1}^{(f)}, \bm\theta_{a}^{(\pi)}, \y)$ {using Alg.~\ref{alg:method_train_alg}}. \label{step:smnn:trans_ud}
                \EndFor
                \State \textbf{return} $\bm\theta_{A}^{(f)}$, $\bm\theta_{A}^{(\pi)}$.
            \end{algorithmic}
        \end{algorithm}
        }
        
        \mname~is described in Alg. \ref{alg:method_outer_alg}.
        We first describe the overall algorithm, given in Alg.~\ref{alg:method_outer_alg} / Fig.~\ref{fig:smn_bd}, and proceed to describe the subordinate algorithms, Alg.~\ref{alg:method_train_alg} / Fig.~\ref{fig:cond_update_bd} and Alg.~\ref{alg:method_update_step} / Fig.~\ref{fig:update_bd} in turn.
        
        In Alg.~\ref{alg:method_outer_alg} / Fig.~\ref{fig:smn_bd}, we begin by initialising the network parameters (lines \ref{step:smnn:init_prop} and \ref{step:smnn:init_trans} of Alg.~\ref{alg:method_outer_alg}).
        It is not crucial how these parameters are initialised, but we note that our software implementation utilises element wise random uniform initialisations for $\A_l$ and $\b_l$, with values in the range $\pm\sqrt{d_l}$, following \citep{kidger2021equinox}.
        We then learn an initial value for the transition kernel (line \ref{step:smnn:bs_trans} of Alg.~\ref{alg:method_outer_alg}) by optimising the network parameters of the transition kernel in a bootstrap particle filter, wherein the transition and proposal distributions are the same.
        
        We then perform $A$ alternating conditional updates. 
        We optimise each of $\bm\theta^{(f)}$ and $\bm\theta^{(\pi)}$ conditional on the other, similar to a coordinate descent method.
        At iteration $a$, we first optimise the proposal distribution parameters $\bm\theta_{a}^{(\pi)}$, conditional on the value of the transition kernel parameters $\bm\theta_{a-1}^{(f)}$.
        We then optimise the transition kernel parameters $\bm\theta_{a}^{(f)}$ conditional on the value of the proposal distribution parameters $\bm\theta_{a}^{(\pi)}$.

            {\footnotesize
            \begin{algorithm}
                \caption{Conditional update algorithm \hfill $\mathrm{ConditionalUpdate}(B, J, \bm\theta_0, \bm\theta_\mathrm{static}, \y)$ \hspace{1em} \null}\label{alg:method_train_alg}
                \begin{algorithmic}[1]
                    \State \textbf{Input:} Number of batches $B$, steps per batch $J$, initial parameters $\bm\theta_0$, static parameters $\bm\theta_{\mathrm{static}}$, observations $\y$.
                    \State Initialise $\bm\theta_{0,J} := \bm\theta_0$. \label{step:cond:init_outer}
                    \For{$b \in 1,\dots,B$}
                    \State Set $\y^{(b)} := \y_{1:\lceil bT/B \rceil}$.
                    \State Set $\bm\theta_{b,0} := \bm\theta_{b-1,J}$.
                        \For{$j \in 1,\dots,J$}
                            \State Set $\bm\theta_{b,j} := \mathrm{UpdateStep}(\bm\theta_{b, j-1}, \bm\theta_{\mathrm{static}}, \y^{(b)})$ using Alg.~\ref{alg:method_update_step}. \label{step:cond:update}
                        \EndFor
                    \EndFor
                    \State \textbf{return} $\bm\theta^{}_{B,J}$.
                \end{algorithmic}
            \end{algorithm}
            }
        We now describe the conditional update, given in Alg.~\ref{alg:method_train_alg} and Fig.~\ref{fig:cond_update_bd}.
        The conditional update algorithm takes an initial value of the parameters of the distribution of interest $\bm\theta_0$, and the parameters of the distribution we are conditioning on $\bm\theta_{static}$.
        We split the observation series $\y_{1:T}$ into $B$ batches, with the $b$-th batch given by $\y^{(b)} = \y_{1:\lceil bT/B \rceil}$.
        Note that this construction implies $\y^{(1)} \subseteq \y^{(2)} \subseteq \cdots \subseteq \y^{(B-1)} \subseteq \y^{(B)}$.
        We construct these telescoping batches of observations to avoid likelihood issues stemming from unadapted parameters, since the first sampled trajectories often have an extremely small log-likelihood, which causes numerical errors when computing the weights in Alg.~\ref{alg:sir_pf}, leaving to the gradients of the log-likelihood exploding \citep{doucet2009tutorial, scibior2021differentiable}.
        
        We iterate over each of the $b$ batches, and for each batch perform $J$ optimisation steps.
        For the $j$-th optimisation step of batch $b$, we run Alg.~\ref{alg:method_update_step} with $\bm\theta_{\mathrm{learn}} = \bm\theta_{b, j-1}, \bm\theta_{\mathrm{static}} = \bm\theta_{\mathrm{static}}$ and observations $\y = \y^{(b)} := \y_{1:\lceil bT/B \rceil}$.
        Note that Alg.~\ref{alg:method_train_alg} / Fig.~\ref{fig:cond_update_bd} runs the particle filter a total of $JB$ times, once for each of the $J$ steps taken in each of the $B$ batches.
        As we perform $2$ runs of Alg.~\ref{alg:method_train_alg} / Fig.~\ref{fig:cond_update_bd} in each of the $A$ iterative steps of Alg.~\ref{alg:method_outer_alg} / Fig.~\ref{fig:smn_bd}, we run the particle filter a total of $2ABJ$ times in total.

            {\footnotesize
            \begin{algorithm}
                \caption{Update step \hfill $\mathrm{UpdateStep}(\bm\theta_\mathrm{learn}, \bm\theta_\mathrm{static}, \y)$ \hspace{1em} \null}\label{alg:method_update_step}
                \begin{algorithmic}[1]
                    \State \textbf{Input:} Parameters to learn $\bm\theta_\mathrm{learn}$, static parameters $\bm\theta_\mathrm{static}$ observations $\y$.
                    
                    \State Set $f(\x_{t}|\x_{t-1}) := \mathcal{GM}(\mathrm{NN}^{(f)}(\x_{t-1};[\bm\theta_\mathrm{learn}, \bm\theta_\mathrm{static}]))$. \label{step:update:make_trans}
                        \State $\pi(\x_t|\x_{t-1}, \y_{t}) := \mathcal{GM}(\mathrm{NN}^{(\pi)}(\x_{t-1}, \y_{t};[\bm\theta_\mathrm{learn}, \bm\theta_\mathrm{static}]))$. \label{step:update:make_prop}
                    \State Run a particle filter (Alg. \ref{alg:sgd_dpf}) with transition kernel $f$, proposal distribution $\pi$, and observations $\y$. \label{step:update:run_pf}
                    \State Obtain $\ell(\bm\theta_{learn})$ and $\nabla \ell(\bm\theta_\mathrm{learn})$ from the particle filter via Eq.~\eqref{eq:loglik} and autodifferentiation. \label{step:update:get_outputs}
                    \State Obtain $\bm\theta_{out}$ by applying a gradient based update to $\bm\theta_\mathrm{learn}$ with gradients $\nabla \ell(\bm\theta_\mathrm{learn})$. \label{step:update:gradient_update}
                    \State \textbf{return} $\bm\theta_{out}$.
                \end{algorithmic}
            \end{algorithm}
            }

        Finally, we describe a single step of the inner update algorithm, given by Alg.~\ref{alg:method_update_step} / Fig.~\ref{fig:update_bd}.
        In each step, we construct the transition kernel (line \ref{step:update:make_trans} of Alg.~\ref{alg:method_update_step}) and proposal distribution (line \ref{step:update:make_prop} of Alg.~\ref{alg:method_update_step}) corresponding to the parameters at that step.
        We construct each distribution using Eq.~\eqref{eq:makedist}, and thereby discard the parameters not relevant to a given network.
        In particular, $\mathrm{NN}^{(\cdot)}(\cdot; [\bm\theta_{\mathrm{learn}}, \bm\theta_{\mathrm{static}}])$ takes as arguments both the static parameter, and the parameter we are learning when constructing both distributions.
        However, each distribution takes either the static parameter, or the parameter we are learning, and we discard the unused parameter for the purposes of constructing that distribution.
        For example, when learning the the transition kernel $f$, $\mathrm{NN}^{(f)}$ takes both $\bm\theta_\mathrm{learn}$, corresponding to $\bm\theta^{(f)}$ and $\bm\theta_\mathrm{static}$, corresponding to $\bm\theta^{(\pi)}$, in line \ref{step:update:make_trans} of Alg.~\ref{alg:method_update_step}, but discards $\bm\theta_\mathrm{static}$ as $\bm\theta_\mathrm{static}$ parametrises the proposal distribution $\pi$ when learning $f$.

        After constructing the transition kernel $f$ and the proposal distribution $\pi$, we run a particle filter with these distributions (line \ref{step:update:run_pf} of Alg.~\ref{alg:method_update_step}), using the observations $\y$, which may be a subset of the overall series of observations.
        From the particle filter we obtain an estimate of the log-likelihood of the parameter $\bm\theta_\mathrm{learn}$, denoted by $\ell\left(\bm\theta_\mathrm{learn}\right)$, and an estimate of the gradient of the log-likelihood with respect to $\bm\theta_\mathrm{learn}$, denoted by $\nabla \ell\left(\bm\theta_\mathrm{learn}\right)$ (line \ref{step:update:get_outputs}).
        We then utilise a gradient update scheme, such as ADAM \citep{kingma2015method} or Novograd \citep{ginsburg2019training, li2019jasper}, to apply a gradient update to the parameter $\bm\theta_\mathrm{learn}$, and output the result of this update to be used within line \ref{step:cond:update} of Alg.~\ref{alg:method_train_alg}.

        {Additionally, we present our algorithm and its constituent sub-algorithms as frame diagrams, so as to visualise the logical flow, as well as the interconnectedness of the method and its sub-methods.
        Fig.~\ref{fig:smn_bd} represents Alg.~\ref{alg:method_outer_alg}.
        Fig.~\ref{fig:cond_update_bd} represents Alg.~\ref{alg:method_train_alg}, and
        Fig.~\ref{fig:update_bd} represents Alg.~\ref{alg:method_update_step},
        }

        \begin{figure}[H]
            \centering
            {
            \scalebox{0.8}{{
\begin{tikzpicture}[auto, node distance=1.5cm,>=latex']
    \node [block_noline, name=inputs] {$[B, J, A, \mathbf y]$};
    
    \node [block_noline, name=outer_inits, right of=inputs, node distance=4.5cm] {$\begin{aligned} \bm\theta_{0}^{(\pi)} &:= \{\mathbf{A}_1^{(\pi)},\mathbf{b}_1^{(\pi)},\dots,\mathbf{A}^{(\pi)}_{L^{(\pi)}},\mathbf{b}^{(\pi)}_{L^{(\pi)}}\}\\\bm\theta_{-1}^{(f)} &:= \{\mathbf{A}_1^{(f)},\mathbf{b}_1^{(f)},\dots,\mathbf{A}^{(f)}_{L^{(f)}},\mathbf{b}^{(f)}_{L^{(f)}}\}\end{aligned}$};
    
    \draw[->] (inputs) -- (outer_inits);

    \node [block, name=bootstrap_dynamics, right of=outer_inits, node distance=6cm]{$\mathrm{ConditionalUpdate}(B, J, \bm\theta_{-1}^{(f)}, \bm\theta_{-1}^{(f)}, \mathbf y)$};

    \draw[->] (outer_inits) -- (bootstrap_dynamics);

    \node [input, name=inner_sep_1, below of=inputs, node distance=1.05cm]{};
    \node [name=init_a, below of=inner_sep_1, node distance=1.05cm]{$a := 1$};
    \draw[->] (bootstrap_dynamics) |- (inner_sep_1) -| (init_a);

    \node [block, name=cu_1, right of=init_a, node distance=4.5cm]{$\begin{aligned}\bm\theta_{a}^{(\pi)} &:= \mathrm{ConditionalUpdate}(B, J, \bm\theta_{a-1}^{(\pi)}, \bm\theta_{a-1}^{(f)}, \mathbf y)\\\bm\theta_{a}^{(f)} &:= \mathrm{ConditionalUpdate}(B, J, \bm\theta_{a-1}^{(f)}, \bm\theta_{a}^{(\pi)}, \mathbf y)\end{aligned}$};
    \draw[->] (init_a) -- node [name=a_loop_left]{} (cu_1);

    \node [output, name=output, right of=cu_1, node distance=6cm] {};
    \node [output, name=output_true, right of=output, node distance=1cm] {};
    \draw[-] (cu_1) -- node [name=a_loop_right]{} (output);
    \draw[->] (output) -- (output_true);
    
    \node [right of=output_true, node distance=0.8cm] {$[\bm\theta_{A}^{(f)}$, $\bm\theta_{A}^{(\pi)}]$} ;

    \node [below of=cu_1, node distance=1cm, name=a_loop_inter]{$a := a+1, a \leq A$};

    \draw[->] (output) |- (a_loop_inter) -| (a_loop_left);;

\end{tikzpicture}
}}
            }
            \caption{Frame diagram of Alg.~\ref{alg:method_outer_alg}, $\mathrm{StateMixNN}(B, J, A, \mathbf y, \mathrm{NN}^{(f)}, \mathrm{NN}^{(\pi)})$.}
            \label{fig:smn_bd}
        \end{figure}

        \begin{figure}[H]
            \centering
            {
            \scalebox{0.8}{\begin{tikzpicture}[auto, node distance=1.5cm,>=latex']
    \node [block_noline, name=input_params] {$[B, J, \bm\theta_0, \bm\theta_{\mathrm{static}}]$};

    \node [block_noline, right of=input_params, name=init_overall, node distance=3cm] {$\begin{aligned} \bm\theta_{0, J} &:= \bm\theta_0,\\b&:=1\end{aligned}$};
    \draw [->] (input_params) -- (init_overall);

    \node [block_noline, right of=init_overall, name=init_inner, node distance=3cm] {$\begin{aligned} \mathbf y^{(b)} &:= \mathbf y_{1:\lceil bT/B \rceil},\\\bm\theta_{b, 0}&:=\bm\theta_{b-1, J},\\j&:=1\end{aligned}$};

    \draw[->] (init_overall) -- node[output, name=b_loop]{} (init_inner);

    \node [block, right of=init_inner, name=update_step, node distance=4cm] {$\mathrm{UpdateStep}(\bm\theta_{b,j-1}, \bm\theta_{\mathrm{static}})$};

    \draw[->] (init_inner) -- node[output, name=j_loop]{} (update_step);

    \node [output, name=output, right of=update_step, node distance=3cm] {};
    \node [right of=output, node distance=0.4cm] {$\bm\theta_{b,j}$} ;
    \draw[->] (update_step) -- node[output, name=j_loop_right]{} (output);

    \node [name=j_loop_inter, above of=update_step, node distance=1cm] {$j = j+1, j \leq J$};
    \node [name=b_loop_inter, above of=init_inner, node distance=1.5cm] {$b = b+1, b \leq B$};
    \draw[->] (j_loop_right.center) |- (j_loop_inter) -| (j_loop.center);
    \draw[->] (j_loop_right.center) |- (b_loop_inter) -| (b_loop.center);
    
\end{tikzpicture}}
            }
            \caption{Frame diagram of Alg.~\ref{alg:method_train_alg}, $\mathrm{ConditionalUpdate}(B, J, \bm\theta_0, \bm\theta_\mathrm{static}, \mathbf y)$.}
            \label{fig:cond_update_bd}
        \end{figure}
    
        \begin{figure}[H]
            \centering
            {
            \scalebox{0.8}{\begin{tikzpicture}[auto, node distance=1.5cm,>=latex']
    \node [block_noline, name=input] {$\begin{aligned}
        f(\mathbf{x}_t|\mathbf{x}_{t-1}; \bm\theta_{\mathrm{learn}} \bm\theta_{\mathrm{static}}),\\
        \pi(\mathbf{x}_t|\mathbf{x}_{t-1}, \mathbf{y}_t; \bm\theta_{\mathrm{learn}}, \bm\theta_{\mathrm{static}}),\\
        \mathbf y
    \end{aligned}$};
    \node [block, right of=input, node distance=4cm] (particle_filter) {$\mathrm{DPF}(\mathbf{y}, f, \pi)$};
    \node [block, right of=particle_filter, node distance=5.5cm] (update) {$\mathrm{Update}(\ell, \nabla \ell)$)};
    \node [output, right of=update, name=output, node distance=3cm] {};

    \draw [->] (input) -- (particle_filter);

    \draw [->] (particle_filter) -- node {$\ell(\bm\theta_{\mathrm{learn}}), \nabla\ell(\bm\theta_{\mathrm{learn}})$} (update);

    \node [right of=output, name=true_output, node distance=0.35cm]{$\bm\theta_{\mathrm{out}}$};
    \draw [->] (update) -- (output);
\end{tikzpicture}}
            }
            \caption{Frame diagram of Alg.~\ref{alg:method_update_step}, $\mathrm{UpdateStep}(\bm\theta_\mathrm{learn}, \bm\theta_\mathrm{static}, \mathbf y)$.}
            \label{fig:update_bd}
        \end{figure}
        
    \subsection{Discussion}

        Our method approximates the transition and proposal distributions by multivariate Gaussian mixture distributions.
        These mixtures are capable of representing complex unknown distributions, and are in many situations both more interpretable and more reliable than methods common in machine learning literature such as normalising flows \citep{gama2022unrolling, gama2023unsupervised, rezende2015variational, trippe2018conditional}. 
        Furthermore, in the context of particle filters, normalising flows have been observed to be susceptible to overfitting, in addition to being less intuitive and harder to train \citep{gama2023unsupervised, gama2022unrolling}.

        We note that the particle weights $w_t$ depend on the samples $\x_{0:t}$, which are drawn from the proposal distribution that we are learning.
        Furthermore, the computation of the weights requires evaluating the density of the transition kernel.
        Therefore, we require a way to propagate gradients through particle resampling and sampling the proposal mixture distribution.
        We use the stop-gradient differentiable particle filter of \cite{scibior2021differentiable}, given in Alg.~\ref{alg:sgd_dpf}, to propagate gradients through the resampling step of the particle filter.
        By including resampling in the training using this method we improve the convergence characteristics of \mname, and allow our method to be utilised to learn complex systems, as weight degeneracy is addressed in training via the resampling step \citep{karkus2018particle,chen2023overview}.

        In order to sample the multivariate Gaussian mixture distributions $f$ and $\pi$, we draw the component from a categorical distribution, and then sample the associated multivariate Gaussian distribution.
        We reparametrise the categorical distribution using the Gumbel-Softmax reparametrisation \citep{jang2016categorical}, thereby allowing gradient propagation through the categorical sampling.
        However, it is equally valid to use a stop-gradient sampling for the categorical distribution, which is computationally more efficient.
        We can propagate gradients through sampling a multivariate Gaussian using the reparametrisation trick \citep{kingma2013auto}, where we externalise the randomness to a input of an affine transformation, which, crucially, is independent $\bm\theta^{(f)}$ and $\bm\theta^{(\pi)}$, and therefore has zero gradient with respect to our learned parameters.
        The above, in combination with the differentiable particle filter method of \cite{scibior2021differentiable}, allow us to compute the gradient of Eq.~\eqref{eq:loglik} with respect to our parameters $\bm\theta^{(p)} = \{\A^{(p)}_1,\b^{(p)}_1,\dots,\A^{(p)}_{L^{(p)}},\b^{(p)}_{L^{(p)}}\}, p \in \{\pi, f\}$, and therefore train the proposal and transition networks using a gradient scheme, such as \citep{kingma2015method, liu2019variance}.

    \section{Discussion about the \mname~framework}
    \label{sec:discuss}
        We now discuss some characteristics of Alg.~\ref{alg:method_outer_alg} / Fig.~\ref{fig:smn_bd}, and explain several of the choices we have made.
        First, we discuss the reasoning for using the multivariate Gaussian mixture distribution for our approximating distributions.
        Then, we justify the restriction to diagonal covariance and equal mixture weights.
        Third, we justify the alternating estimation scheme of Alg.\ref{alg:method_outer_alg}.
        Finally, we discuss how we combat the phenomena of likelihood degeneracy and concentration in our method.

    \subsection{Choice of a Gaussian mixture distribution for $\pi$ and $f$, and extensions}
        In this work, we propose in Eq.~\eqref{eq:makedist} to use a multivariate Gaussian mixture distribution to approximate the optimal proposal distribution and the state transition kernel.
        The Gaussian mixture is able to approximate a wide range of distributions, whilst retaining computational speed.
        Furthermore, Gaussian mixtures are interpretable, and are easier to understand and infer from than approximating distributions such as normalising flows \cite{tabak2013family}.

        Most deep learning frameworks have fast, differentiable implementations of the Gaussian distribution, and typically will have the capability to combine distributions resulting in a mixture distribution.
        Using standard methods such as these allows for rapid, optimised implementation of our method in a given deep learning framework, for example PyTorch \cite{paszke2019torch}, JAX \cite{jax2018github}, and Tensorflow \cite{tensorflow2015whitepaper} are popular frameworks, which all have standard constructs which can be used to construct efficient multivariate Gaussian mixtures.

        We note that our method can be trivially extended to mixture distributions for which the components can be parametrised by their location and scale parameters, with an example of such an extension being a mixture of multivariate t distributions with a fixed degree of freedom $\nu$. 
        To do so, we would replace the $\mathcal{N}(\bm\mu^{(s)},\C^{(s)})$ mixands in Eq.~\eqref{eq:makedist} with $\mathit{t}_{\nu}(\bm\mu^{(s)},\C^{(s)})$, noting that $\nu$ is fixed.
        Using the t distribution may capture tail behaviour better; however, the t distribution is more expensive both to sample from and to evaluate the density of.
        For example, the density of the multivariate t distribution requires evaluating the $\Gamma$ function, whereas the density for the Gaussian distribution requires only exponentiation and standard linear algebra operations.

    \subsection{Restriction to diagonal covariance and equal weights in the approximating mixture}
    
        In Section~\ref{sec:arch} we present our method, restricting the covariance of mixture components, $\C^{(s)}$, to a diagonal form, and the mixture weights to be uniformly equal to $S^{-1}$ in Eq.~\eqref{eq:makedist}. 
        We present our reasoning for these restrictions below.

    \noindent\textbf{Covariances}
        As we estimate both the state and proposal distributions $f$ and $\pi$, we infer complex dependencies between state dimensions in the interaction between the distributions, and therefore do not need to estimate full covariance matrices. 
        We assume the form of the covariance matrices in order to reduce the number of parameters required; we can parametrise a $N$ dimensional distribution by $2N$ parameters, whereas estimating the full covariance matrix would require $N + N(N+1)/2$ parameters, $N$ for the mean $\bm\mu^{(s)}$, and $N(N+1)/2$ for the covariance $\C^{(s)}$, as covariance matrices are positive semi-definite and are therefore symmetric.
        Furthermore, increasing the number of estimated parameters also increases the required dimension of the output of each network, $\z_L$, and therefore increases the dimension of the hidden layers parameters, $\A_l$ and $\b_l$, $l \in 1, \dots, L$, required to obtain a comparable estimation power, drastically increasing the number of network parameters that must be learnt at each step.
        
        A diagonal covariance also allows for a efficient implementation of the multivariate Gaussian distribution.
        For example, we do not need to compute a full matrix inverse ${\C^{(s)}}^{-1}$ when computing the density of a multivariate Gaussian distribution; if we know that the covariance $\C^{(s)}$ is diagonal, we can simply invert the diagonal element-wise, as ${\C^{(s)}}^{-1} = \mathrm{diag}(1 / \c^{(s)})$ for diagonal $\C^{(s)}$.

    \noindent\textbf{Mixture weights}
        In Eq.~\eqref{eq:makedist} we impose that the mixture weights are uniformly equal to $S^{-1}$, where $S$ is the number of mixture components.
        We can loosen this restriction and learn the mixture weights, but this is a difficult task.
        Much of this difficulty comes from the fact that changes in the log-likelihood can now be brought about by changing either the component parameters $\bm\mu^{(s)}$ and $\C^{(s)}$, or the mixture weights, which we denote here by $m^{(s)}$.
        
        For each mixture component, the mean parameter $\bm\mu^{(s)}$ and the covariance parameter $\C^{(s)}$ are vector valued and matrix valued, although we restrict the covariance to be diagonal as Eq.~\eqref{eq:makedist}. 
        The mean and covariance of each component are thus determined by $2 d_x$ values; however, the mixture weight $m^{(s)}$ is a single value, which has a far larger impact on the likelihood than any single element of the mean or covariance.
        Therefore, the gradient of the log-likelihood, computed via Eq.~\eqref{eq:loglik} with backpropagation, with respect to the network parameters $\bm\theta^{(\pi)}$ and $\bm\theta^{(f)}$, will be primarily attributable to the effect these parameters have on the mixture weight, which will result in convergence issues as only the mixture weights will meaningfully change between iterations. 
        Identifiability would also be a problem, as changes in the likelihood could be attributed to changes in either the mean/covariance, or changes to the mixture weights.
        
        The parameters would thus be significantly more difficult to train, and the method given in Alg.~\ref{alg:method_outer_alg} / Fig.~\ref{fig:smn_bd} would not be sufficient.
        A potential solution would be to introduce two more networks, each outputting only the mixture weights for the transition and proposal distributions respectively, and then to train these networks as part of the iterative step of Alg.~\ref{alg:method_train_alg} / Fig.~\ref{fig:cond_update_bd}, thereby training 4 networks in an alternating scheme. 
        For the sake of brevity, and for the clarity of this paper, we utilise the equal weighted mixture, which we note can well approximate arbitrarily weighted mixtures with a sufficiently large number of components.

    \subsection{Use of an alternating scheme when estimating network parameters}
        In order to learn the network parameters $\bm\theta^{(f)}$ and $\bm\theta^{(\pi)}$, Alg.~\ref{alg:method_outer_alg} / Fig.~\ref{fig:smn_bd} uses an alternating scheme to learn each parameter conditional on the value of the other.
        Learning one parameter conditional on the other stabilises inference, as the parameters heavily influence each other given that the proposal and transition interact in the weighting step of Alg.~\ref{alg:sgd_dpf}.
        Learning in such a manner may lead to identifiability issues, as any change in the weights or particles, and hence in the log-likelihood, can be attributed to either a change in the transition kernel, or a change in the proposal distribution.
        Furthermore, changes in one distribution should be reflected by changes in the other distribution, as both are tightly linked in interpretation and usage within the filter.
        
        However, if we were to update networks $\mathrm{NN}^{(f)}$ and $\mathrm{NN}^{(\pi)}$ simultaneously, i.e., we replace Alg.~\ref{alg:method_train_alg} / Fig.~\ref{fig:cond_update_bd} with a gradient update of both $\bm\theta^{(f)}$ and $\bm\theta^{(\pi)}$ at the same time, we cannot attribute changes in the likelihood to a specific network, nor can we update each network conditional on the changes made to the other network.
        In addition, learning both networks at the same time can lead to divergence, as each network may change in such a way that they yield numerically incompatible distributions; in such distributions the transition kernel has near zero probability mass where the proposal has large probability mass, thereby leading to numerically zero weights when computed in line \ref{step_dpf:weights} of Alg.~\ref{alg:sgd_dpf}.
     
        In Alg.~\ref{alg:method_outer_alg} / Fig.~\ref{fig:smn_bd} we learn each distribution conditional on the other, and therefore do not suffer from an identifiability problem, as any changes are attributable only to the learned distribution.
        Further, we learn the effects the changes made to each distribution have on the other, hence stabilising the learning, as the distributions do not adapt at the same time.
        Alternating between learning and conditioning on each distribution allows the distribution to adapt to each other, and mitigates the potential problem of distributions diverging in density.

        {The order of the alternating scheme does not matter asymptotically, and we obtain nearly identical results when we learn $\bm\theta^{(f)}$ in each and then learn $\bm\theta^{(\pi)}$ in each iteration, instead of learning $\bm\theta^{(\pi)}$ and then $\bm\theta^{(f)}$, however, more iterations of the scheme are required.
        As we initialise $\bm\theta^{(f)}$ using an estimate based on the bootstrap particle filter, learning $\bm\theta^{(\pi)}$ first in the first iteration, we have a better value for $\bm\theta^{(\pi)}$ when we learn $\bm\theta^{(f)}$ conditional on $\bm\theta^{(\pi)}$.
        This improves the convergence of the algorithm, and reduces the number of iterations required as both parameters are initially learnt conditioned on reasonable estimates of the other parameter.}
        
    \subsection{Combating likelihood degeneracy when estimating parameters using the particle filter}
        Our method learns $\bm\theta^{(f)}$ and $\bm\theta^{(\pi)}$ by maximising the estimated log likelihood, $\ell([\bm\theta^{(f)}, \bm\theta^{(\pi)}])$, using an alternating scheme.
        We estimate the log-likelihood using Eq.~\eqref{eq:loglik}, which, in turn, uses the particle weights $w_t^{(k)}$, $t = 1,\dots,T$, $k = 1,\dots,K$, from all iterations of the filter.
        Therefore, the success of our method hinges on the accurate and stable computation of these weights, which is a challenge in particle filtering, as the weight $w_t^{(k)}$ of a given particle $\x_t^{(k)}$ can be very close to zero.
        One way to numerically stabilise weight computations is to use log weights \cite{sarkka2013bayesian}. 
        Implementing log weights is a simple change, as we need only rewrite the weight calculation in line \ref{step_dpf:weights} of Alg.~\ref{alg:sgd_dpf} to use log likelihoods, and rewrite the normalisation to remain on the log scale.

        We define $\mathrm{logsumexp}([x_1, x_2, \dots, x_N]) := \log\left(\sum_{n=1}^N \exp(x_n)\right)$.\footnote{
        The $\mathrm{logsumexp}$ can be additionally stabilised by observing that $\mathrm{logsumexp}([x_n]_{n=1}^{N}) = \max([x_n]_{n=1}^{N}) + \mathrm{logsumexp}([x_n - \max([x_m]_{m=1}^{N})]_{n=1}^{N})$, which helps to avoid numerical overflow when evaluating $\mathrm{logsumexp}$. 
        Note that this modification is typically already present in pre-existing implementations of $\mathrm{logsumexp}$ such as those present in \cite{paszke2019torch, jax2018github, tensorflow2015whitepaper}.}
        Note that 
            \begin{equation}
                \begin{alignedat}{2}
                 \log\left(w_t^{(k)}\right) &= \log\left(\frac{g(\y_t|\x_t^{(k)}; \bm\theta^{(g)})f(\x_t^{(k)}|\x_{t-1}^{a_{t}^{(k)}}; \bm\theta^{(f)})}{\pi(\x_t|\x_{t-1}^{a_{t}^{(k)}}, \y_{t}; \bm\theta^{(\pi)})}\right),\\
                 &= \log\left(g\left(\y_t|\x_t^{(k)}; \bm\theta^{(g)}\right)\right) + \log\left(f\left(\x_t^{(k)}|\x_{t-1}^{a_{t}^{(k)}}; \bm\theta^{(f)}\right)\right) - \log\left(\pi\left(\x_t|\x_{t-1}^{a_{t}^{(k)}}, \y_{t}; \bm\theta^{(\pi)}\right)\right).
                \end{alignedat}
            \end{equation}
        By writing $p_t^{(k)} = \tilde{w}_{t-1}^{(k)}w_t^{(k)}$, and noting that $\log\left(p_t^{(k)}\right) = \log\left(\tilde{w}_{t-1}^{(k)}\right) + \log\left(w_t^{(k)}\right)$, we have
            \begin{equation}
                \begin{alignedat}{2}
                    \bar{w}_t^{(k)} &= \frac{p_t^{(k)}}{\sum_{k=1}^{K}p_t^{(k)}} = \frac{\exp\left(\log\left(p_t^{(k)}\right)\right)}{\sum_{k=1}^{K}\exp\left(\log\left(p_t^{(k)}\right)\right)},\\
                    \log\left(\bar{w}_t^{(k)}\right) &= \log\left(p_t^{(k)}\right) - \log\left(\sum_{k=1}^{K}\exp\left(\log\left(p_t^{(k)}\right)\right)\right),\\
                    &= \log\left(p_t^{(k)}\right) - \mathrm{logsumexp}\left(\left[\log\left(p_t^{(k)}\right)\right]_{k=1}^{K}\right),\\
                    &= \log\left(\tilde{w}_{t-1}^{(k)}\right) + \log\left(w_t^{(k)}\right) - \mathrm{logsumexp}\left(\left[\log\left(\tilde{w}_{t-1}^{(k)}\right) + \log\left(w_t^{(k)}\right)\right]_{k=1}^{K}\right),
                \end{alignedat}
            \end{equation}
        and can construct the log-likelihood estimator following Eq.~\eqref{eq:loglik} directly using the log weights. 
        The use of log weights significantly improves numerical stability; nevertheless, it does not address the issue of likelihood concentration, nor does it mitigate the problem of learning parameters with initial value having very low likelihood.
 
        State-space models in general suffer from likelihood concentration, where $\ell(\bm\theta|\y_{1:T})$, which we estimate by Eq.~\eqref{eq:loglik}, becomes increasingly concentrated around a single value of $\bm\theta$ as $T$ increases.
        Likelihood concentration thereby results in vanishingly small likelihoods for parameter values that are not well adapted to the system, leading to the gradient of the parameter values being numerically zero, and the parameter is hence unable to be learnt, without careful initialisation near a value with high likelihood.
        The issue is therefore that we must choose an initial parameter that has high likelihood, but we assume we do not know the parameter beforehand;  these statements contradict each other.

        We address likelihood concentration, and hence the problem of learning parameters under random initialisation, using observation batching.
        We infer our parameters, $\bm\theta^{(f)}$ and $\bm\theta^{(\pi)}$, in Alg.~\ref{alg:method_update_step} using $B$ increasingly large batches $\y^{(b)}$, $b \in \{1, \dots, B\}$, of the observation series, where $\y^{(1)} \subseteq \y^{(2)} \subseteq \cdots \subseteq \y^{(B-1)} \subseteq \y^{(B)}$, choosing $\y^{(B)} := \y$.
        Warming up the $\bm\theta^{(f)}$ and $\bm\theta^{(\pi)}$ parameters before learning on the entire series mitigates the effect of likelihood concentration for unadapted parameters, and, when combined with log weights, allows our method to be used on long observation series for complex problems.

        We initially learn our parameters, $\bm\theta^{(f)}$ and $\bm\theta^{(\pi)}$, on a small subset, $\y^{(1)}$, of the observation series $\y$.
        Thereby, we obtain a relatively diffuse parameter likelihood function compared to that obtained with a longer observation series.
        Therefore, we can initialise the parameters of the neural networks, $\A_l, \ \b_l, \ l \in 1, \dots, L$, randomly, as the likelihood and its gradient will not be numerically zero for unadapted parameters, due to the relatively diffuse parameter likelihood function.
        Further, under the above method, our estimated $f$ and $\pi$ distributions incorporate information from the observation series $\y$ in sequence, adapting to one batch of observations $\y^{(b)}$ before taking in more. 
        Sequentially incorporating observation information into our estimation prevents behaviour where the start and end of the series are well represented by the learnt parameters, but the middle is not; this is a common problem when using neural networks to learn between-step dynamics in time series models \citep{chen2018neural, yan2019robustness, rackauckas2019diffeqflux}.        

    \section{Numerical Experiments}
    \label{sec:experiments}

    We will now illustrate the performance of our proposed method on two systems. 
    First, we will test our method on a non-linear polynomial system: the Lorenz 96 chaotic oscillator.
    We will then apply our method to a non-linear non-polynomial system: the Kuramoto oscillator.
    In both instances we compare our method to the \acrfull{iapf} \citep{elvira2018improved} and the \acrfull{bpf} \citep{Gordon93}, {as well as StateMixNN \cite{cox2024end}, which uses similar techniques but learns only the proposal distribution.}
    {The IAPF utilises measurement information to improve the proposal distribution in a pre-weighting step, aiming to improve the characteristics of the proposal distribution. }
    Note that all of the methods that we compare against require more information than \mname, as the transition kernel must be known.
    Finally, we note that all methods are run using the same differentiable particle filter, given in Alg.~\ref{alg:sgd_dpf}, which does not modify the forward pass behaviour of the SIR particle filter.    

    \subsection{Lorenz 96 model}
        We consider a stochastic version of the Lorenz 96 model \citep{lorenz1996predictability}, a dynamical system known to exhibit chaotic behaviour.
        We insert additive noise terms to obtain a stochastic system to use for testing, and discretise using the Euler-Maruyama scheme, resulting in the system
            \begin{align}
                \label{eq:l93_ssm}
                \begin{split}
                    x_{i, t+1} &= x_{i, t} + \Delta t (x_{i-1, t}(x_{i+1, t} - x_{i-2, t}) - x_{i, t} + F) + \sqrt{\Delta t} \cdot v_{i,t+1},\\
                    y_{i, t+1} &= x_{i, t+1} + \sqrt{\Delta t} \cdot r_{i,t+1},
                \end{split}
            \end{align}
        for $i \in \{1,\dots,d_x\}$ and $t \in \{0,\dots,T\}$, where we define $x_{-1} := x_{d_x - 1}, x_0 := x_{d_x},$ and $x_{d_x+1} := x_1$, $v_t \sim \mathcal{N}(\bm 0, \bm\Sigma_v)$, $r_t \sim \mathcal{N}(\bm 0, \bm\Sigma_r)$, and $F$ is a forcing constant; we use $F = 8$.

        We set $\Delta t = 0.05$ in Eq.~\eqref{eq:l93_ssm}. 
        Unless explicitly stated otherwise, we choose the dimension of the system as $d_x = d_y = 20$, and set 
        $\bm\Sigma_v= 0.25\boldsymbol{I}_{d_x}$ and $\bm\Sigma_r= 0.1\boldsymbol{I}_{d_x}$.
        We initialise the hidden state at $\x_0$ such that $x_{1, 0} = 1$, with all other elements equal to $0$.
        We assess the proposed method in terms of relative improvement in \gls{mse}, showing the accuracy of the method as a fraction of the \gls{mse} obtained with the \gls{bpf} \citep{Gordon93}. 
        {The relative improvement in MSE is given by  
        \begin{equation}
            \mathrm{RI_{MSE}} = \frac{\mathrm{MSE}_\mathrm{method}}{\mathrm{MSE}_\mathrm{baseline}},
        \end{equation}
        where $\mathrm{MSE}_\mathrm{method}$ is the MSE of the method we are testing, and $\mathrm{MSE}_\mathrm{baseline}$ is the MSE of our baseline method in the same task.
        The \gls{mse} compares the weighted mean of the samples (the estimated state) with the true underlying hidden state. 
        For this model metric, lower is better, as it indicates a lower MSE than the baseline.
        We use the bootstrap particle filter as our baseline method in all instances.
        Note that the \gls{mse} is not the optimisation target of our method.}
        
        We compare our method with the \gls{iapf} \citep{elvira2018improved}, to illustrate the performance of a standard improved proposal {utilising observation information}.
        {Note that there do exist other methods that utilise observation information to attempt to improve the proposal, such as the Gaussian particle filter \cite{kotecha2003gaussian} and the auxiliary particle filter \cite{pitt1999filtering}, however of those tested the IAPF performed best on our problem set.}
        {Further, we compare our method to PropMixNN \cite{cox2024end}, the method from which~\mname~is extended, which we fix at $S=6$ components, and use the parameters given in the originating paper.
        Note that the \gls{iapf}, the \gls{bpf}, and PropMixNN, all require the state transition model to be known, and therefore cannot be applied to a general system as~\mname~can.}
        {We note that all methods tested are implemented using the same stop-gradient DPF \cite{scibior2021differentiable} used by~\mname, however, no parameters are learned for the \gls{iapf} and the \gls{bpf}.}
        We compute the \gls{mse} for 200 independent runs of the filter, and plot the mean and symmetric $95\%$ intervals.
    
        We test the proposed method using a variable number of mixture components, with $S \in \{1,6,10\}$. 
        All variants utilise the same network architecture, with 3 layers of output sizes $d_1 = 128, d_2 = 256, d_3 = 2Sd_x$ for both networks.
        For the activation function $\rho_l$ in Eq.~\eqref{eq:nn_setup}, we set $\rho_{1:2}(x) = \mathrm{relu}(x) = \max(0, x)$, and $\rho_{3}(x) = x$, with this applying to both the proposal and transition networks.
        We train \mname~using the ADAM optimiser \citep{kingma2015method}, using a fixed learning rate of $3\cdot10^{-3}$, and setting the parameters of Alg.~\ref{alg:method_outer_alg} / Fig.~\ref{fig:smn_bd} to $B=\lceil T/5 \rceil, J=50, A=20$.
        We train the method using a series of observations distinct from those on which we test \mname; however, all series are instances of the Lorenz 96 system.

        \noindent\textbf{Variable number of particles.}
        We test \mname~for a variable number of particles $K$, with $K \in \{30, 50, 100, 200\}$. 
        We use a fixed series length $T=100$.
        {We present the results in Fig. \ref{fig:variableparticles}, which shows that 
        \mname~outperforms the \gls{bpf} for all given values of $K$, obtaining at most $0.9$ times the \gls{mse} of the \gls{bpf}, and typically less than $0.75$ the \gls{mse}.
        \mname~with $S=10$ components suffers with few particles, performing worse than the other parametrisations of \mname, as few samples are taken from each component, therefore the gradient estimates are less reliable, as the gradients relating to each component are dependent on only a few evaluations. 
        In the case of $S=6$ there are more evaluations of each component than in the case of $S=10$, which allows for sufficient information to learn the transition dynamics.}
        Our method outperforms the \gls{iapf} at all tested numbers of particles, by an increasingly large margin as the number of particles increases.
            \begin{figure}[H]
                    \centering
                    \begin{tikzpicture}
\begin{axis}[
    xlabel={\footnotesize Number of Particles},
    ylabel={\footnotesize $RI_{MSE}$},
    xmin=25, xmax=205,
    ymin=0.25, ymax=1.1,
    xtick={30,50,100,200},
    ytick={0, 0.25, 0.5, 0.75, 1},
    legend style={at={(0.5,-0.3)},anchor=north,font=\footnotesize},
    ymajorgrids=true,
    grid style=dashed,
    ticklabel style = {font = \footnotesize},
    height = 1.6in,
    width = 0.90\textwidth,
]

\addplot [color=black,dashed] coordinates {(0,1) (500,1)};
\addplot [name path=upper10comps_100ts,draw=none] table[x=x,y expr=\thisrow{y}+\thisrow{err}] {10comps_100ts.dat};
\addplot [name path=lower10comps_100ts,draw=none] table[x=x,y expr=\thisrow{y}-\thisrow{err}] {10comps_100ts.dat};
\addplot[
    color=green,
    mark=o,
    ]
    table[x=x, y=y]{10comps_100ts.dat};
\addplot [fill=green!50, fill opacity=0.5] fill between[of=upper10comps_100ts and lower10comps_100ts];
\addplot [name path=upper6comps_100ts,draw=none] table[x=x,y expr=\thisrow{y}+\thisrow{err}] {6comps_100ts.dat};
\addplot [name path=lower6comps_100ts,draw=none] table[x=x,y expr=\thisrow{y}-\thisrow{err}] {6comps_100ts.dat};
\addplot[
    color=blue,
    mark=diamond,
    ]
    table[x=x, y=y]{6comps_100ts.dat};
\addplot [fill=blue!50, fill opacity=0.5] fill between[of=upper6comps_100ts and lower6comps_100ts];
\addplot [name path=upper1comps_100ts,draw=none] table[x=x,y expr=\thisrow{y}+\thisrow{err}] {1comps_100ts.dat};
\addplot [name path=lower1comps_100ts,draw=none] table[x=x,y expr=\thisrow{y}-\thisrow{err}] {1comps_100ts.dat};
\addplot[
    color=red,
    mark=square,
    ]
    table[x=x, y=y]{1comps_100ts.dat};
\addplot [fill=red!50, fill opacity=0.5] fill between[of=upper1comps_100ts and lower1comps_100ts];
\addplot [name path=upperAPF_100ts,draw=none] table[x=x,y expr=\thisrow{y}+\thisrow{err}] {APF_100ts.dat};
\addplot [name path=lowerAPF_100ts,draw=none] table[x=x,y expr=\thisrow{y}-\thisrow{err}] {APF_100ts.dat};
\addplot[
    color=orange,
    mark=x,
    ]
    table[x=x, y=y]{APF_100ts.dat};
\addplot [fill=orange!50, fill opacity=0.5] fill between[of=upperAPF_100ts and lowerAPF_100ts];
\addplot [name path=upper6comps_100ts_propmix,draw=none] table[x=x,y expr=\thisrow{y}+\thisrow{err}] {6comps_100ts_propmix.dat};
\addplot [name path=lower6comps_100ts_propmix,draw=none] table[x=x,y expr=\thisrow{y}-\thisrow{err}] {6comps_100ts_propmix.dat};
\addplot[
    color=purple,
    mark=star,
    ]
    table[x=x, y=y]{6comps_100ts_propmix.dat};
\addplot [fill=purple!50, fill opacity=0.5] fill between[of=upper6comps_100ts_propmix and lower6comps_100ts_propmix];

\end{axis}
\end{tikzpicture}
                    \begin{tikzpicture}
\begin{axis}[
    legend style={font=\footnotesize, nodes={scale=0.85, transform shape}},
    hide axis,
    xmin=10,
    xmax=50,
    ymin=0,
    ymax=0.4,
    grid style=dashed,
    legend entries={10 components, 6 components, 1 component, IAPF, PropMixNN (6 components), BPF},
    legend columns = 3,
    height = 1.6in,
    width = 1.6in,
]
\addlegendimage{mark=o, color = green}
\addlegendimage{mark=diamond, color = blue}
\addlegendimage{mark=square, color = red}
\addlegendimage{mark=x, color = orange}
\addlegendimage{mark=star, color = purple}
\addlegendimage{color = black, dashed}
\end{axis}
\end{tikzpicture}
                    \caption{Comparison of StateMixNN with the BPF, IAPF, and PropMixNN over variable numbers of particles. The lines denote mean performance, with bands denoting symmetric $95\%$ intervals.}
                    \label{fig:variableparticles}
            \end{figure}

    \noindent\textbf{Variable series length.}
        Next, we test \mname~with a variable series length $T$, with $T \in \{30, 60, 100, 200, 500\}$. 
        In this case we fix the number of particles $K=100$.
        We show in Fig.~\ref{fig:variablelength} that
        the proposed method obtains lower values of \gls{mse} than the \gls{bpf} and \gls{iapf} for all given values of $T$.
        The $S=10$ component method slightly outperforms the $S=6$ component method, which significantly outperforms the $S=1$ component method.
            \begin{figure}[H]
                \centering
                \begin{tikzpicture}
\begin{axis}[
    xlabel={\footnotesize Series length},
    ylabel={\footnotesize $RI_{MSE}$},
    xmin=20, xmax=510,
    ymin=0.25, ymax=1.1,
    xtick={30,60,100,200,500},
    ytick={0, 0.25, 0.5, 0.75, 1},
    legend style={at={(0.5,-0.3)},anchor=north,font=\footnotesize},
    ymajorgrids=true,
    grid style=dashed,
    ticklabel style = {font = \footnotesize},
    height = 1.6in,
    width = 0.90\textwidth,
]

\addplot [color=black,dashed] coordinates {(0,1) (550,1)};
\addplot [name path=upper10comps_vts,draw=none] table[x=x,y expr=\thisrow{y}+\thisrow{err}] {10comps_vts.dat};
\addplot [name path=lower10comps_vts,draw=none] table[x=x,y expr=\thisrow{y}-\thisrow{err}] {10comps_vts.dat};
\addplot[
    color=green,
    mark=o,
    ]
    table[x=x, y=y]{10comps_vts.dat};
\addplot [fill=green!50, fill opacity=0.5] fill between[of=upper10comps_vts and lower10comps_vts];
\addplot [name path=upper6comps_vts,draw=none] table[x=x,y expr=\thisrow{y}+\thisrow{err}] {6comps_vts.dat};
\addplot [name path=lower6comps_vts,draw=none] table[x=x,y expr=\thisrow{y}-\thisrow{err}] {6comps_vts.dat};
\addplot[
    color=blue,
    mark=diamond,
    ]
    table[x=x, y=y]{6comps_vts.dat};
\addplot [fill=blue!50, fill opacity=0.5] fill between[of=upper6comps_vts and lower6comps_vts];
\addplot [name path=upper1comps_vts,draw=none] table[x=x,y expr=\thisrow{y}+\thisrow{err}] {1comps_vts.dat};
\addplot [name path=lower1comps_vts,draw=none] table[x=x,y expr=\thisrow{y}-\thisrow{err}] {1comps_vts.dat};
\addplot[
    color=red,
    mark=square,
    ]
    table[x=x, y=y]{1comps_vts.dat};
\addplot [fill=red!50, fill opacity=0.5] fill between[of=upper1comps_vts and lower1comps_vts];
\addplot [name path=upperAPF_vts,draw=none] table[x=x,y expr=\thisrow{y}+\thisrow{err}] {APF_vts.dat};
\addplot [name path=lowerAPF_vts,draw=none] table[x=x,y expr=\thisrow{y}-\thisrow{err}] {APF_vts.dat};
\addplot[
    color=orange,
    mark=x,
    ]
    table[x=x, y=y]{APF_vts.dat};
\addplot [fill=orange!50, fill opacity=0.5] fill between[of=upperAPF_vts and lowerAPF_vts];
\addplot [name path=upper6comps_vts_propmix,draw=none] table[x=x,y expr=\thisrow{y}+\thisrow{err}] {6comps_vts_propmix.dat};
\addplot [name path=lower6comps_vts_propmix,draw=none] table[x=x,y expr=\thisrow{y}-\thisrow{err}] {6comps_vts_propmix.dat};
\addplot[
    color=purple,
    mark=star,
    ]
    table[x=x, y=y]{6comps_vts_propmix.dat};
\addplot [fill=purple!50, fill opacity=0.5] fill between[of=upper6comps_vts_propmix and lower6comps_vts_propmix];

\end{axis}
\end{tikzpicture}
                \begin{tikzpicture}
\begin{axis}[
    legend style={font=\footnotesize, nodes={scale=0.85, transform shape}},
    hide axis,
    xmin=10,
    xmax=50,
    ymin=0,
    ymax=0.4,
    grid style=dashed,
    legend entries={10 components, 6 components, 1 component, IAPF, PropMixNN (6 components), BPF},
    legend columns = 3,
    height = 1.6in,
    width = 1.6in,
]
\addlegendimage{mark=o, color = green}
\addlegendimage{mark=diamond, color = blue}
\addlegendimage{mark=square, color = red}
\addlegendimage{mark=x, color = orange}
\addlegendimage{mark=star, color = purple}
\addlegendimage{color = black, dashed}
\end{axis}
\end{tikzpicture}
                \caption{Comparison of StateMixNN with the BPF, IAPF, and PropMixNN over variable series length. The lines denote mean performance, with bands denoting symmetric $95\%$ intervals.}
                \label{fig:variablelength}
            \end{figure}

    \noindent\textbf{Variable state noise.}
        We now test \mname~for a variable state noise $\bm\Sigma_v= \sigma_v^2\boldsymbol{I}_{20}$, with $\sigma_v^2 \in \{0.05, 0.1, 0.25, 0.5, 1\}$. 
        In this case, we fix the number of particles $K=100$ and the series length $T=100$.
        Fig. \ref{fig:variablenoise} shows that
        \mname~is superior to the \gls{bpf} for all given values of $\sigma_v^2$.
        The improvement in accuracy is lesser for small noise variances, as small perturbations give a concentrated distribution of the state values at the next time step.
        However, the performance improves for larger values of $\sigma_v^2$. 
        This is due to the chaotic behaviour of the system, which leads to multimodal distributions for the next state, as the state follows one of several diverging paths.
        The mixture in the proposed method captures this behaviour, with components representing different modes, thereby outperforming both the \gls{bpf} and the \gls{iapf}.  
            \begin{figure}[H]
                \centering
                \begin{tikzpicture}
\begin{axis}[
    xlabel={\footnotesize $\sigma_v^2$},
    ylabel={\footnotesize $RI_{MSE}$},
    xmin=0.04, xmax=1.01,
    ymin=0.25, ymax=1.1,
    xtick={0.05, 0.25, 0.5, 1},
    ytick={0, 0.25, 0.5, 0.75, 1},
    legend style={at={(0.5,-0.3)},anchor=north,font=\footnotesize},
    ymajorgrids=true,
    grid style=dashed,
    xticklabel style = {/pgf/number format/fixed, /pgf/number format/precision = 2},
    ticklabel style = {font = \footnotesize},
    height = 1.6in,
    width = 0.90\textwidth,
]

\addlegendimage{mark=o, color = green}
\addlegendimage{mark=diamond, color = blue}
\addlegendimage{mark=square, color = red}
\addplot [color=black, dashed] coordinates {(0,1) (1.1,1)};

\addplot [name path=upper10comps_varnoise,draw=none] table[x=x,y expr=\thisrow{y}+\thisrow{err}] {10comps_varnoise.dat};
\addplot [name path=lower10comps_varnoise,draw=none] table[x=x,y expr=\thisrow{y}-\thisrow{err}] {10comps_varnoise.dat};
\addplot[
    color=green,
    mark=o,
    ]
    table[x=x, y=y]{10comps_varnoise.dat};
\addplot [fill=green!50, fill opacity=0.5] fill between[of=upper10comps_varnoise and lower10comps_varnoise];
\addplot [name path=upper6comps_varnoise,draw=none] table[x=x,y expr=\thisrow{y}+\thisrow{err}] {6comps_varnoise.dat};
\addplot [name path=lower6comps_varnoise,draw=none] table[x=x,y expr=\thisrow{y}-\thisrow{err}] {6comps_varnoise.dat};
\addplot[
    color=blue,
    mark=diamond,
    ]
    table[x=x, y=y]{6comps_varnoise.dat};
\addplot [fill=blue!50, fill opacity=0.5] fill between[of=upper6comps_varnoise and lower6comps_varnoise];
\addplot [name path=upper1comps_varnoise,draw=none] table[x=x,y expr=\thisrow{y}+\thisrow{err}] {1comps_varnoise.dat};
\addplot [name path=lower1comps_varnoise,draw=none] table[x=x,y expr=\thisrow{y}-\thisrow{err}] {1comps_varnoise.dat};
\addplot[
    color=red,
    mark=square,
    ]
    table[x=x, y=y]{1comps_varnoise.dat};
\addplot [fill=red!50, fill opacity=0.5] fill between[of=upper1comps_varnoise and lower1comps_varnoise];
\addplot [name path=upperAPF_varnoise,draw=none] table[x=x,y expr=\thisrow{y}+\thisrow{err}] {APF_varnoise.dat};
\addplot [name path=lowerAPF_varnoise,draw=none] table[x=x,y expr=\thisrow{y}-\thisrow{err}] {APF_varnoise.dat};
\addplot[
    color=orange,
    mark=x,
    ]
    table[x=x, y=y]{APF_varnoise.dat};
\addplot [fill=orange!50, fill opacity=0.5] fill between[of=upperAPF_varnoise and lowerAPF_varnoise];
\addplot [name path=upper6comps_varnoise_propmix,draw=none] table[x=x,y expr=\thisrow{y}+\thisrow{err}] {6comps_varnoise_propmix.dat};
\addplot [name path=lower6comps_varnoise_propmix,draw=none] table[x=x,y expr=\thisrow{y}-\thisrow{err}] {6comps_varnoise_propmix.dat};
\addplot[
    color=purple,
    mark=star,
    ]
    table[x=x, y=y]{6comps_varnoise_propmix.dat};
\addplot [fill=purple!50, fill opacity=0.5] fill between[of=upper6comps_varnoise_propmix and lower6comps_varnoise_propmix];

\end{axis}
\end{tikzpicture}
                \begin{tikzpicture}
\begin{axis}[
    legend style={font=\footnotesize, nodes={scale=0.85, transform shape}},
    hide axis,
    xmin=10,
    xmax=50,
    ymin=0,
    ymax=0.4,
    grid style=dashed,
    legend entries={10 components, 6 components, 1 component, IAPF, PropMixNN (6 components), BPF},
    legend columns = 3,
    height = 1.6in,
    width = 1.6in,
]
\addlegendimage{mark=o, color = green}
\addlegendimage{mark=diamond, color = blue}
\addlegendimage{mark=square, color = red}
\addlegendimage{mark=x, color = orange}
\addlegendimage{mark=star, color = purple}
\addlegendimage{color = black, dashed}
\end{axis}
\end{tikzpicture}
                \caption{Comparison of StateMixNN with the BPF, IAPF, and PropMixNN over variable state noise magnitude. The lines denote mean performance, with bands denoting symmetric $95\%$ intervals.}
                \label{fig:variablenoise}
            \end{figure}
        
        \noindent\textbf{Variable system dimension.}
        Finally, we test \mname~over variable state and observation dimension $d_x=d_y$, with $d_x \in \{5, 10, 20, 40, 60\}$. 
        We fix the number of particles $K=100$, and the series length $T=100$.
        Fig.~\ref{fig:variabledimension} shows that
        \mname~is superior to the \gls{bpf} for all given values of $d_x$.
        We observe that the margin of out-performance decreases as the state dimension increases, but seems to stabilise after $d_x = 40$.
        This is due to the static number of iterations used in training, as the state-space is easier to learn for smaller $d_x$, and therefore requires fewer iterations to achieve the same level of performance.
        We observe that \mname~performs well at all tested values of $d_x$, and does not rapidly deteriorate in performance when increasing $d_x$ without changing the training regime.
        Furthermore, the filters with a larger number of components in the learned distributions outperform those with a smaller number of components at all times, displaying the mixture distributions ability to approximate complex systems.
            \begin{figure}[H]
                \centering
                
                \begin{tikzpicture}
\begin{axis}[
    xlabel={\footnotesize $d_x$},
    ylabel={\footnotesize $RI_{MSE}$},
    xmin=4, xmax=61,
    ymin=0.25, ymax=1.1,
    xtick={5, 10, 20, 40, 60},
    ytick={0, 0.25, 0.5, 0.75, 1},
    legend style={at={(0.5,-0.3)},anchor=north,font=\footnotesize},
    ymajorgrids=true,
    grid style=dashed,
    xticklabel style = {/pgf/number format/fixed, /pgf/number format/precision = 2},
    ticklabel style = {font = \footnotesize},
    height = 1.6in,
    width = 0.90\textwidth,
]

\addlegendimage{mark=o, color = green}
\addlegendimage{mark=diamond, color = blue}
\addlegendimage{mark=square, color = red}
\addplot [color=black, dashed] coordinates {(0,1) (1.1,1)};

\addplot [name path=upper10comps_vardim,draw=none] table[x=x,y expr=\thisrow{y}+\thisrow{err}] {10comps_vardim.dat};
\addplot [name path=lower10comps_vardim,draw=none] table[x=x,y expr=\thisrow{y}-\thisrow{err}] {10comps_vardim.dat};
\addplot[
    color=green,
    mark=o,
    ]
    table[x=x, y=y]{10comps_vardim.dat};
\addplot [fill=green!50, fill opacity=0.5] fill between[of=upper10comps_vardim and lower10comps_vardim];
\addplot [name path=upper6comps_vardim,draw=none] table[x=x,y expr=\thisrow{y}+\thisrow{err}] {6comps_vardim.dat};
\addplot [name path=lower6comps_vardim,draw=none] table[x=x,y expr=\thisrow{y}-\thisrow{err}] {6comps_vardim.dat};
\addplot[
    color=blue,
    mark=diamond,
    ]
    table[x=x, y=y]{6comps_vardim.dat};
\addplot [fill=blue!50, fill opacity=0.5] fill between[of=upper6comps_vardim and lower6comps_vardim];
\addplot [name path=upper1comps_vardim,draw=none] table[x=x,y expr=\thisrow{y}+\thisrow{err}] {1comps_vardim.dat};
\addplot [name path=lower1comps_vardim,draw=none] table[x=x,y expr=\thisrow{y}-\thisrow{err}] {1comps_vardim.dat};
\addplot[
    color=red,
    mark=square,
    ]
    table[x=x, y=y]{1comps_vardim.dat};
\addplot [fill=red!50, fill opacity=0.5] fill between[of=upper1comps_vardim and lower1comps_vardim];
\addplot [name path=upperAPF_vardim,draw=none] table[x=x,y expr=\thisrow{y}+\thisrow{err}] {APF_vardim.dat};
\addplot [name path=lowerAPF_vardim,draw=none] table[x=x,y expr=\thisrow{y}-\thisrow{err}] {APF_vardim.dat};
\addplot[
    color=orange,
    mark=x,
    ]
    table[x=x, y=y]{APF_vardim.dat};
\addplot [fill=orange!50, fill opacity=0.5] fill between[of=upperAPF_vardim and lowerAPF_vardim];
\addplot [name path=upper6comps_vardim_propmix,draw=none] table[x=x,y expr=\thisrow{y}+\thisrow{err}] {6comps_vardim_propmix.dat};
\addplot [name path=lower6comps_vardim_propmix,draw=none] table[x=x,y expr=\thisrow{y}-\thisrow{err}] {6comps_vardim_propmix.dat};
\addplot[
    color=purple,
    mark=star,
    ]
    table[x=x, y=y]{6comps_vardim_propmix.dat};
\addplot [fill=purple!50, fill opacity=0.5] fill between[of=upper6comps_vardim_propmix and lower6comps_vardim_propmix];

\end{axis}
\end{tikzpicture}
                \begin{tikzpicture}
\begin{axis}[
    legend style={font=\footnotesize, nodes={scale=0.85, transform shape}},
    hide axis,
    xmin=10,
    xmax=50,
    ymin=0,
    ymax=0.4,
    grid style=dashed,
    legend entries={10 components, 6 components, 1 component, IAPF, PropMixNN (6 components), BPF},
    legend columns = 3,
    height = 1.6in,
    width = 1.6in,
]
\addlegendimage{mark=o, color = green}
\addlegendimage{mark=diamond, color = blue}
\addlegendimage{mark=square, color = red}
\addlegendimage{mark=x, color = orange}
\addlegendimage{mark=star, color = purple}
\addlegendimage{color = black, dashed}
\end{axis}
\end{tikzpicture}
                \caption{Comparison of StateMixNN with the BPF, IAPF, and PropMixNN over variable system dimension. The lines denote mean performance, with bands denoting symmetric $95\%$ intervals.}
                \label{fig:variabledimension}
            \end{figure}

    \subsection{Kuramoto oscillator}
        The Kuramoto oscillator \citep{kuramoto1984chemical} is a mathematical model that describes the behavior of a system of $d_x$ phase-coupled oscillators. 
        The model is described by
        {
            \begin{align}
                \begin{split}
                \label{eq:kuramoto_eq}
                    \frac{\mathrm{d}\theta_i}{\mathrm{d}t} = \omega_i + d_x^{-1}\sum_{j=1}^{d_x}C\sin(\theta_j - \theta_i),
                \end{split}
            \end{align}
        }
        where $\theta_i$ denotes the phase of the $i$th oscillator, and $C \in \reals$ is the coupling constant between oscillators.
        This does not restrict $\bm\theta$ however, which will, in general, diverge to an infinity as $t \rightarrow \infty$.
        To address this, we transform Eq.~\eqref{eq:kuramoto_eq} by introducing derived parameters $R \in \reals^+$ and {$\phi \in [-\pi, \pi]$} such that
        {
            \begin{align}
                \begin{split}
                \label{eq:kuramoto_eq_imag_reparam}
                    R\exp(\sqrt{-1}\phi) &= d_x^{-1} \sum_{j=1}^{d_x} \exp\left(\sqrt{-1}\theta_j\right),\\
                    \frac{\mathrm{d}\theta_i}{\mathrm{d}t} &= \omega_i + CR \sin(\phi - \theta_i),
                \end{split}
            \end{align}
        }
        which restricts $\bm\theta \in [-\pi, \pi]^{d_x}$.
        We insert additive Gaussian noise to Eq.~\eqref{eq:kuramoto_eq_imag_reparam}, and discretise using the Euler-Maruyama scheme, yielding the NLSSM
            \begin{equation}
                \begin{alignedat}{2}
                    R\exp(\sqrt{-1}\phi) &= d_x^{-1} \sum_{j=1}^{d_x} \exp\left(\sqrt{-1}x_{j, t}\right),\\
                    x_{i, t+1} &= x_{i, t} + \Delta t \left(\omega_i + CR\sin(\phi - x_i)\right) + \sqrt{\Delta t} \cdot v_{i, t+1}, \\
                    y_{i, t+1} &= x_{i, t+1} + \sqrt{\Delta t} \cdot r_{i, t+1},
                \end{alignedat}
            \end{equation}
        for $i \in \{1, \dots, d_x\}$.
        We choose $d_x = 20$, and {$C = 0.8$}.
        We set $\bm\Sigma_v = \sigma_v^2 \boldsymbol{I}_{d_x}, \bm\Sigma_r = \sigma_r^2 \boldsymbol{I}_{d_x}$, with {$\sigma_v = 0.1, \sigma_r = 0.005$} unless stated otherwise, which tests the performance of our method in the case that the observation is much more informative than the state, where the standard \gls{bpf} is known to suffer.
        We discretise this model with a time step of $\Delta t = 0.05$.
        We sample $\omega_i \sim \mathcal{N}(0.5, 0.5^2)$, and $\x_{i,0} \sim U(-\pi, \pi)$.
        We run the system until $t=10$, and then begin collecting observations.

    \noindent\textbf{Variable series length.}
        {
        We test \mname~with a variable series length $T$, with $T \in \{30, 60, 100, 200, 500\}$ on the Kuramoto oscillator
        In this case we fix the number of particles $K=100$.
        We show in Fig.~\ref{fig:variablelength_kura} that the proposed method obtains lower values of \gls{mse} than the \gls{bpf} and \gls{iapf} for all given values of $T$.
        The $S=10$ component method outperforms the $S=6$ component method, which in turn outperforms the $S=1$ component method.        
            \begin{figure}[H]
                \centering
                \begin{tikzpicture}
\begin{axis}[
    xlabel={\footnotesize Series length},
    ylabel={\footnotesize $RI_{MSE}$},
    xmin=20, xmax=510,
    ymin=0.0, ymax=1.1,
    xtick={30,60,100,200,500},
    ytick={0, 0.25, 0.5, 0.75, 1},
    legend style={at={(0.5,-0.3)},anchor=north,font=\footnotesize},
    ymajorgrids=true,
    grid style=dashed,
    ticklabel style = {font = \footnotesize},
    height = 1.6in,
    width = 0.90\textwidth,
]

\addplot [color=black,dashed] coordinates {(0,1) (550,1)};
\addplot [name path=upper10comps_vts_kura,draw=none] table[x=x,y expr=\thisrow{y}+\thisrow{err}] {10comps_vts_kura.dat};
\addplot [name path=lower10comps_vts_kura,draw=none] table[x=x,y expr=\thisrow{y}-\thisrow{err}] {10comps_vts_kura.dat};
\addplot[
    color=green,
    mark=o,
    ]
    table[x=x, y=y]{10comps_vts_kura.dat};
\addplot [fill=green!50, fill opacity=0.5] fill between[of=upper10comps_vts_kura and lower10comps_vts_kura];
\addplot [name path=upper6comps_vts_kura,draw=none] table[x=x,y expr=\thisrow{y}+\thisrow{err}] {6comps_vts_kura.dat};
\addplot [name path=lower6comps_vts_kura,draw=none] table[x=x,y expr=\thisrow{y}-\thisrow{err}] {6comps_vts_kura.dat};
\addplot[
    color=blue,
    mark=diamond,
    ]
    table[x=x, y=y]{6comps_vts_kura.dat};
\addplot [fill=blue!50, fill opacity=0.5] fill between[of=upper6comps_vts_kura and lower6comps_vts_kura];
\addplot [name path=upper1comps_vts_kura,draw=none] table[x=x,y expr=\thisrow{y}+\thisrow{err}] {1comps_vts_kura.dat};
\addplot [name path=lower1comps_vts_kura,draw=none] table[x=x,y expr=\thisrow{y}-\thisrow{err}] {1comps_vts_kura.dat};
\addplot[
    color=red,
    mark=square,
    ]
    table[x=x, y=y]{1comps_vts_kura.dat};
\addplot [fill=red!50, fill opacity=0.5] fill between[of=upper1comps_vts_kura and lower1comps_vts_kura];
\addplot [name path=upperAPF_vts_kura,draw=none] table[x=x,y expr=\thisrow{y}+\thisrow{err}] {APF_vts_kura.dat};
\addplot [name path=lowerAPF_vts_kura,draw=none] table[x=x,y expr=\thisrow{y}-\thisrow{err}] {APF_vts_kura.dat};
\addplot[
    color=orange,
    mark=x,
    ]
    table[x=x, y=y]{APF_vts_kura.dat};
\addplot [fill=orange!50, fill opacity=0.5] fill between[of=upperAPF_vts_kura and lowerAPF_vts_kura];
\addplot [name path=upper6comps_vts_kura_propmix,draw=none] table[x=x,y expr=\thisrow{y}+\thisrow{err}] {6comps_vts_kura_propmix.dat};
\addplot [name path=lower6comps_vts_kura_propmix,draw=none] table[x=x,y expr=\thisrow{y}-\thisrow{err}] {6comps_vts_kura_propmix.dat};
\addplot[
    color=purple,
    mark=star,
    ]
    table[x=x, y=y]{6comps_vts_kura_propmix.dat};
\addplot [fill=purple!50, fill opacity=0.5] fill between[of=upper6comps_vts_kura_propmix and lower6comps_vts_kura_propmix];
\end{axis}
\end{tikzpicture}
                \begin{tikzpicture}
\begin{axis}[
    legend style={font=\footnotesize, nodes={scale=0.85, transform shape}},
    hide axis,
    xmin=10,
    xmax=50,
    ymin=0,
    ymax=0.4,
    grid style=dashed,
    legend entries={10 components, 6 components, 1 component, IAPF, PropMixNN (6 components), BPF},
    legend columns = 3,
    height = 1.6in,
    width = 1.6in,
]
\addlegendimage{mark=o, color = green}
\addlegendimage{mark=diamond, color = blue}
\addlegendimage{mark=square, color = red}
\addlegendimage{mark=x, color = orange}
\addlegendimage{mark=star, color = purple}
\addlegendimage{color = black, dashed}
\end{axis}
\end{tikzpicture}
                \caption{Comparison of StateMixNN with the BPF, IAPF, and PropMixNN over variable series length on the Kuramoto oscillator. The lines denote mean performance, with bands denoting symmetric $95\%$ intervals.}
                \label{fig:variablelength_kura}
            \end{figure}
        }

    \noindent\textbf{Variable number of particles.}
        We test \mname~for a variable number of particles $K$, with $K \in \{30, 50, 100, 200\}$. 
        We use a fixed series length $T=100$.
        {We present the results in Fig. \ref{fig:variableparticles_kura}, which shows that 
        \mname~outperforms the \gls{bpf} and \gls{iapf} for all given values of $K$.
        \mname~with $S=10$ components suffers with few particles, performing worse than the other parametrisations of \mname, as few samples are taken from each component, therefore the gradient estimates are more dispersed in the $S=10$ case, making training less reliable. }
            \begin{figure}[H]
                    \centering
                    \begin{tikzpicture}
\begin{axis}[
    xlabel={\footnotesize Number of Particles},
    ylabel={\footnotesize $RI_{MSE}$},
    xmin=25, xmax=205,
    ymin=0.0, ymax=1.1,
    xtick={30,50,100,200},
    ytick={0, 0.25, 0.5, 0.75, 1},
    legend style={at={(0.5,-0.3)},anchor=north,font=\footnotesize},
    ymajorgrids=true,
    grid style=dashed,
    ticklabel style = {font = \footnotesize},
    height = 1.6in,
    width = 0.90\textwidth,
]

\addplot [color=black,dashed] coordinates {(0,1) (500,1)};
\addplot [name path=upper10comps_K_kura,draw=none] table[x=x,y expr=\thisrow{y}+\thisrow{err}] {10comps_K_kura.dat};
\addplot [name path=lower10comps_K_kura,draw=none] table[x=x,y expr=\thisrow{y}-\thisrow{err}] {10comps_K_kura.dat};
\addplot[
    color=green,
    mark=o,
    ]
    table[x=x, y=y]{10comps_K_kura.dat};
\addplot [fill=green!50, fill opacity=0.5] fill between[of=upper10comps_K_kura and lower10comps_K_kura];
\addplot [name path=upper6comps_K_kura,draw=none] table[x=x,y expr=\thisrow{y}+\thisrow{err}] {6comps_K_kura.dat};
\addplot [name path=lower6comps_K_kura,draw=none] table[x=x,y expr=\thisrow{y}-\thisrow{err}] {6comps_K_kura.dat};
\addplot[
    color=blue,
    mark=diamond,
    ]
    table[x=x, y=y]{6comps_K_kura.dat};
\addplot [fill=blue!50, fill opacity=0.5] fill between[of=upper6comps_K_kura and lower6comps_K_kura];
\addplot [name path=upper1comps_K_kura,draw=none] table[x=x,y expr=\thisrow{y}+\thisrow{err}] {1comps_K_kura.dat};
\addplot [name path=lower1comps_K_kura,draw=none] table[x=x,y expr=\thisrow{y}-\thisrow{err}] {1comps_K_kura.dat};
\addplot[
    color=red,
    mark=square,
    ]
    table[x=x, y=y]{1comps_K_kura.dat};
\addplot [fill=red!50, fill opacity=0.5] fill between[of=upper1comps_K_kura and lower1comps_K_kura];
\addplot [name path=upperAPF_K_kura,draw=none] table[x=x,y expr=\thisrow{y}+\thisrow{err}] {APF_K_kura.dat};
\addplot [name path=lowerAPF_K_kura,draw=none] table[x=x,y expr=\thisrow{y}-\thisrow{err}] {APF_K_kura.dat};
\addplot[
    color=orange,
    mark=x,
    ]
    table[x=x, y=y]{APF_K_kura.dat};
\addplot [fill=orange!50, fill opacity=0.5] fill between[of=upperAPF_K_kura and lowerAPF_K_kura];
\addplot [name path=upper6comps_K_kura_propmix,draw=none] table[x=x,y expr=\thisrow{y}+\thisrow{err}] {6comps_K_kura_propmix.dat};
\addplot [name path=lower6comps_K_kura_propmix,draw=none] table[x=x,y expr=\thisrow{y}-\thisrow{err}] {6comps_K_kura_propmix.dat};
\addplot[
    color=purple,
    mark=star,
    ]
    table[x=x, y=y]{6comps_K_kura_propmix.dat};
\addplot [fill=purple!50, fill opacity=0.5] fill between[of=upper6comps_K_kura_propmix and lower6comps_K_kura_propmix];
\end{axis}
\end{tikzpicture}
                    \begin{tikzpicture}
\begin{axis}[
    legend style={font=\footnotesize, nodes={scale=0.85, transform shape}},
    hide axis,
    xmin=10,
    xmax=50,
    ymin=0,
    ymax=0.4,
    grid style=dashed,
    legend entries={10 components, 6 components, 1 component, IAPF, PropMixNN (6 components), BPF},
    legend columns = 3,
    height = 1.6in,
    width = 1.6in,
]
\addlegendimage{mark=o, color = green}
\addlegendimage{mark=diamond, color = blue}
\addlegendimage{mark=square, color = red}
\addlegendimage{mark=x, color = orange}
\addlegendimage{mark=star, color = purple}
\addlegendimage{color = black, dashed}
\end{axis}
\end{tikzpicture}
                    \caption{Comparison of StateMixNN with the BPF, IAPF, and PropMixNN over variable number of particles on the Kuramoto oscillator. The lines denote mean performance, with bands denoting symmetric $95\%$ intervals.}
                    \label{fig:variableparticles_kura}
            \end{figure}

    {\noindent\textbf{Variable state noise.}
        We now test \mname~on the Kuramoto oscillator with variable state noise $\bm\Sigma_v= \sigma_v^2\boldsymbol{I}_{d_x}$, with $\sigma_v \in \{0.025, 0.05, 0.1, 0.25, 0.5\}$. 
        In this case, we fix the number of particles $K=100$ and the series length $T=100$.
        }

        {
        We observe in Fig.~\ref{fig:variablenoise_kura} that the performance remains consistent over state noise magnitudes when compared to the performance of the bootstrap particle filter. 
        We observe an apparent decrease in performance for very low state noise magnitude for all tested methods, which is explained by the bootstrap particle filter performing better the closer $\sigma_v$ is to  $\sigma_r$.
            \begin{figure}[H]
                \centering
                \begin{tikzpicture}
\begin{axis}[
    xlabel={\footnotesize $\sigma_v$},
    ylabel={\footnotesize $RI_{MSE}$},
    xmin=0.04, xmax=0.51,
    ymin=0.20, ymax=1.1,
    xtick={0.05, 0.1, 0.25, 0.5},
    ytick={0.25, 0.5, 0.75, 1},
    legend style={at={(0.5,-0.3)},anchor=north,font=\footnotesize},
    ymajorgrids=true,
    grid style=dashed,
    ticklabel style = {font = \footnotesize},
    xticklabel style={
  /pgf/number format/precision=3,
  /pgf/number format/fixed},
    height = 1.6in,
    width = 0.90\textwidth,
]

\addplot [color=black,dashed] coordinates {(0,1) (1,1)};
\addplot [name path=upper10comps_vsigma_kura,draw=none] table[x=x,y expr=\thisrow{y}+\thisrow{err}] {10comps_vsigma_kura.dat};
\addplot [name path=lower10comps_vsigma_kura,draw=none] table[x=x,y expr=\thisrow{y}-\thisrow{err}] {10comps_vsigma_kura.dat};
\addplot[
    color=green,
    mark=o,
    ]
    table[x=x, y=y]{10comps_vsigma_kura.dat};
\addplot [fill=green!50, fill opacity=0.5] fill between[of=upper10comps_vsigma_kura and lower10comps_vsigma_kura];
\addplot [name path=upper6comps_vsigma_kura,draw=none] table[x=x,y expr=\thisrow{y}+\thisrow{err}] {6comps_vsigma_kura.dat};
\addplot [name path=lower6comps_vsigma_kura,draw=none] table[x=x,y expr=\thisrow{y}-\thisrow{err}] {6comps_vsigma_kura.dat};
\addplot[
    color=blue,
    mark=diamond,
    ]
    table[x=x, y=y]{6comps_vsigma_kura.dat};
\addplot [fill=blue!50, fill opacity=0.5] fill between[of=upper6comps_vsigma_kura and lower6comps_vsigma_kura];
\addplot [name path=upper1comps_vsigma_kura,draw=none] table[x=x,y expr=\thisrow{y}+\thisrow{err}] {1comps_vsigma_kura.dat};
\addplot [name path=lower1comps_vsigma_kura,draw=none] table[x=x,y expr=\thisrow{y}-\thisrow{err}] {1comps_vsigma_kura.dat};
\addplot[
    color=red,
    mark=square,
    ]
    table[x=x, y=y]{1comps_vsigma_kura.dat};
\addplot [fill=red!50, fill opacity=0.5] fill between[of=upper1comps_vsigma_kura and lower1comps_vsigma_kura];
\addplot [name path=upperAPF_vsigma_kura,draw=none] table[x=x,y expr=\thisrow{y}+\thisrow{err}] {APF_vsigma_kura.dat};
\addplot [name path=lowerAPF_vsigma_kura,draw=none] table[x=x,y expr=\thisrow{y}-\thisrow{err}] {APF_vsigma_kura.dat};
\addplot[
    color=orange,
    mark=x,
    ]
    table[x=x, y=y]{APF_vsigma_kura.dat};
\addplot [fill=orange!50, fill opacity=0.5] fill between[of=upperAPF_vsigma_kura and lowerAPF_vsigma_kura];
\addplot [name path=upper6comps_vsigma_kura_propmix,draw=none] table[x=x,y expr=\thisrow{y}+\thisrow{err}] {6comps_vsigma_kura_propmix.dat};
\addplot [name path=lower6comps_vsigma_kura_propmix,draw=none] table[x=x,y expr=\thisrow{y}-\thisrow{err}] {6comps_vsigma_kura_propmix.dat};
\addplot[
    color=purple,
    mark=star,
    ]
    table[x=x, y=y]{6comps_vsigma_kura_propmix.dat};
\addplot [fill=purple!50, fill opacity=0.5] fill between[of=upper6comps_vsigma_kura_propmix and lower6comps_vsigma_kura_propmix];
\end{axis}
\end{tikzpicture}
                \begin{tikzpicture}
\begin{axis}[
    legend style={font=\footnotesize, nodes={scale=0.85, transform shape}},
    hide axis,
    xmin=10,
    xmax=50,
    ymin=0,
    ymax=0.4,
    grid style=dashed,
    legend entries={10 components, 6 components, 1 component, IAPF, PropMixNN (6 components), BPF},
    legend columns = 3,
    height = 1.6in,
    width = 1.6in,
]
\addlegendimage{mark=o, color = green}
\addlegendimage{mark=diamond, color = blue}
\addlegendimage{mark=square, color = red}
\addlegendimage{mark=x, color = orange}
\addlegendimage{mark=star, color = purple}
\addlegendimage{color = black, dashed}
\end{axis}
\end{tikzpicture}
                \caption{Comparison of StateMixNN with the BPF, IAPF, and PropMixNN over variable state noise magnitude. The lines denote mean performance, with bands denoting symmetric $95\%$ intervals.}
                \label{fig:variablenoise_kura}
            \end{figure}
        }

    {\noindent\textbf{Variable system dimension.}
        Finally, we test \mname~over variable state and observation dimension $d_x=d_y$, with $d_x \in \{5, 10, 20, 40, 60\}$. 
        We fix the number of particles $K=100$, and the series length $T=100$.
        We present the results in Fig.~\ref{fig:variabledimension_kura}.
        }
        
        {
        We observe that \mname~is superior to the \gls{bpf} for all given values of $d_x$.
        Furthermore, the margin of out-performance is approximately constant after $d_x = 20$, with \mname~outperforming standard methods by a larger margin for smaller values of $d_x$.
        This is likely due both to a smaller number of parameters being learnt, and less complex transition and proposal distributions being required due to the reduced dimensionality of the problem.
        We observe that \mname~performs well at all tested values of $d_x$, and does not rapidly deteriorate in performance when increasing $d_x$ without changing the training regime.
        Furthermore, the filters with a larger number of components in the learned distributions outperform those with a smaller number of components at all times, displaying the mixture distributions ability to approximate complex systems.
            \begin{figure}[H]
                \centering
                \begin{tikzpicture}
\begin{axis}[
    xlabel={\footnotesize $d_x$},
    ylabel={\footnotesize $RI_{MSE}$},
    xmin=4, xmax=61,
    ymin=0.25, ymax=1.1,
    xtick={5, 10, 20, 40, 60},
    ytick={0, 0.25, 0.5, 0.75, 1},
    legend style={at={(0.5,-0.3)},anchor=north,font=\footnotesize},
    ymajorgrids=true,
    grid style=dashed,
    xticklabel style = {/pgf/number format/fixed, /pgf/number format/precision = 2},
    ticklabel style = {font = \footnotesize},
    height = 1.6in,
    width = 0.90\textwidth,
]

\addlegendimage{mark=o, color = green}
\addlegendimage{mark=diamond, color = blue}
\addlegendimage{mark=square, color = red}
\addplot [color=black, dashed] coordinates {(0,1) (1.1,1)};

\addplot [name path=upper10comps_vardim_kura,draw=none] table[x=x,y expr=\thisrow{y}+\thisrow{err}] {10comps_vardim_kura.dat};
\addplot [name path=lower10comps_vardim_kura,draw=none] table[x=x,y expr=\thisrow{y}-\thisrow{err}] {10comps_vardim_kura.dat};
\addplot[
    color=green,
    mark=o,
    ]
    table[x=x, y=y]{10comps_vardim_kura.dat};
\addplot [fill=green!50, fill opacity=0.5] fill between[of=upper10comps_vardim_kura and lower10comps_vardim_kura];
\addplot [name path=upper6comps_vardim_kura,draw=none] table[x=x,y expr=\thisrow{y}+\thisrow{err}] {6comps_vardim_kura.dat};
\addplot [name path=lower6comps_vardim_kura,draw=none] table[x=x,y expr=\thisrow{y}-\thisrow{err}] {6comps_vardim_kura.dat};
\addplot[
    color=blue,
    mark=diamond,
    ]
    table[x=x, y=y]{6comps_vardim_kura.dat};
\addplot [fill=blue!50, fill opacity=0.5] fill between[of=upper6comps_vardim_kura and lower6comps_vardim_kura];
\addplot [name path=upper1comps_vardim_kura,draw=none] table[x=x,y expr=\thisrow{y}+\thisrow{err}] {1comps_vardim_kura.dat};
\addplot [name path=lower1comps_vardim_kura,draw=none] table[x=x,y expr=\thisrow{y}-\thisrow{err}] {1comps_vardim_kura.dat};
\addplot[
    color=red,
    mark=square,
    ]
    table[x=x, y=y]{1comps_vardim_kura.dat};
\addplot [fill=red!50, fill opacity=0.5] fill between[of=upper1comps_vardim_kura and lower1comps_vardim_kura];
\addplot [name path=upperAPF_vardim_kura,draw=none] table[x=x,y expr=\thisrow{y}+\thisrow{err}] {APF_vardim_kura.dat};
\addplot [name path=lowerAPF_vardim_kura,draw=none] table[x=x,y expr=\thisrow{y}-\thisrow{err}] {APF_vardim_kura.dat};
\addplot[
    color=orange,
    mark=x,
    ]
    table[x=x, y=y]{APF_vardim_kura.dat};
\addplot [fill=orange!50, fill opacity=0.5] fill between[of=upperAPF_vardim_kura and lowerAPF_vardim_kura];
\addplot [name path=upper6comps_vardim_kura_propmix,draw=none] table[x=x,y expr=\thisrow{y}+\thisrow{err}] {6comps_vardim_kura_propmix.dat};
\addplot [name path=lower6comps_vardim_kura_propmix,draw=none] table[x=x,y expr=\thisrow{y}-\thisrow{err}] {6comps_vardim_kura_propmix.dat};
\addplot[
    color=purple,
    mark=star,
    ]
    table[x=x, y=y]{6comps_vardim_kura_propmix.dat};
\addplot [fill=purple!50, fill opacity=0.5] fill between[of=upper6comps_vardim_kura_propmix and lower6comps_vardim_kura_propmix];

\end{axis}
\end{tikzpicture}
                \begin{tikzpicture}
\begin{axis}[
    legend style={font=\footnotesize, nodes={scale=0.85, transform shape}},
    hide axis,
    xmin=10,
    xmax=50,
    ymin=0,
    ymax=0.4,
    grid style=dashed,
    legend entries={10 components, 6 components, 1 component, IAPF, PropMixNN (6 components), BPF},
    legend columns = 3,
    height = 1.6in,
    width = 1.6in,
]
\addlegendimage{mark=o, color = green}
\addlegendimage{mark=diamond, color = blue}
\addlegendimage{mark=square, color = red}
\addlegendimage{mark=x, color = orange}
\addlegendimage{mark=star, color = purple}
\addlegendimage{color = black, dashed}
\end{axis}
\end{tikzpicture}
                \caption{Comparison of StateMixNN with the BPF, IAPF, and PropMixNN over variable system dimension. The lines denote mean performance, with bands denoting symmetric $95\%$ intervals.}
                \label{fig:variabledimension_kura}
            \end{figure}
            }
            
    \section{Conclusion}
        \label{sec:conclusions}
        This work proposes a novel method, called \mname, which simultaneously learns the transition and proposal distributions of a particle filter.
        We utilise a pair of multivariate Gaussian mixture distributions to approximate the transition and proposal distributions, with the means and covariances of the mixands given by the output of a dense neural network.
        {\mname~uses a standard neural network architecture to produce its results, and could potentially be extended or modified to use a different architecture, such as a graph neural network to estimate state connectivity, or a recurrent neural network for non-Markovian systems. }
        
       {We present numerical results for a stochastic Lorenz 96 model, which has highly chaotic behaviour, and for the Kuramoto oscillator, which has a restricted state-space and cannot be expressed using polynomial terms.
        We observe that our method outperforms the bootstrap particle filter, a standard method, as well as the improved auxiliary particle filter, a state-of-the-art method for improving the proposal distribution, both of which require more information about the underlying system than \mname.
        Furthermore, our method here performs similarly to PropMixNN, a method that uses a similar neural proposal, but requires the state model to be known.
        The proposed method does not require knowledge of the hidden state, as we optimise the observation likelihood, which requires only the observations to be known.}
        {The ability of \mname~to operate without knowledge of the hidden state dynamics makes it particularly well-suited for real-world applications where the underlying model is unknown.}
    
    \bibliography{bibliography}
\end{document}